\documentclass[11pt]{article}

\usepackage[preprint]{acl}

\usepackage{times,etoc}
\usepackage{inconsolata}

\usepackage{latexsym,amsmath}
\usepackage{booktabs}
\usepackage{multirow,pifont}
\usepackage{array,hyperref}
\usepackage{graphicx}
\usepackage[table]{xcolor}
\definecolor{lightgreen}{RGB}{144, 238, 144}
   \usepackage{graphicx}
   \usepackage{subcaption}
\usepackage{pifont}

\usepackage[T1]{fontenc}

\usepackage[utf8]{inputenc}
\usepackage{xcolor,mdframed}
  \definecolor{xmlcol}{RGB}{173,214,241}        
  \definecolor{jsoncol}{RGB}{180,230,180}       
  \definecolor{loosecol}{RGB}{255,235,160}      
  \definecolor{freecol}{RGB}{250,190,190}       
  \definecolor{logitcol}{RGB}{210,210,210}      

  \mdfdefinestyle{xmlbox}{backgroundcolor=xmlcol!40,linecolor=xmlcol,linewidth=0.8pt,
    innerleftmargin=6pt,innerrightmargin=6pt,innertopmargin=4pt,innerbottommargin=4pt,
    skipabove=4pt,skipbelow=4pt}
  \mdfdefinestyle{jsonbox}{backgroundcolor=jsoncol!40,linecolor=jsoncol,linewidth=0.8pt,
    innerleftmargin=6pt,innerrightmargin=6pt,innertopmargin=4pt,innerbottommargin=4pt,
    skipabove=4pt,skipbelow=4pt}
  \mdfdefinestyle{loosebox}{backgroundcolor=loosecol!40,linecolor=loosecol,linewidth=0.8pt,
    innerleftmargin=6pt,innerrightmargin=6pt,innertopmargin=4pt,innerbottommargin=4pt,
    skipabove=4pt,skipbelow=4pt}
  \mdfdefinestyle{freebox}{backgroundcolor=freecol!40,linecolor=freecol,linewidth=0.8pt,
    innerleftmargin=6pt,innerrightmargin=6pt,innertopmargin=4pt,innerbottommargin=4pt,
    skipabove=4pt,skipbelow=4pt}
  \mdfdefinestyle{logitbox}{backgroundcolor=logitcol!40,linecolor=logitcol,linewidth=0.8pt,
    innerleftmargin=6pt,innerrightmargin=6pt,innertopmargin=4pt,innerbottommargin=4pt,
    skipabove=4pt,skipbelow=4pt}

\usepackage{microtype,booktabs,amssymb}
\usepackage[most]{tcolorbox}

\newtcolorbox{takeaway}{
  enhanced, breakable,
  colback=white,
  frame hidden,
  borderline west={2pt}{0pt}{black!70},
  left=10pt, right=4pt, top=4pt, bottom=4pt,
  fontupper=\itshape,
}
\usepackage{inconsolata}
\usepackage{booktabs}
\usepackage{multirow}
\usepackage{array}
\usepackage[table]{xcolor}
\usepackage{colortbl}

\definecolor{besthl}{HTML}{E8F5E9}      
\definecolor{rowalt}{HTML}{FAFAFA}      
\definecolor{stdgray}{HTML}{8A8A8A}     

\newcommand{\stddev}[1]{{\scriptsize\textcolor{stdgray}{$\pm$#1}}}
\newcommand{\best}[2]{\cellcolor{besthl}\textbf{#1}\,\stddev{#2}}
\newcommand{\cell}[2]{#1\,\stddev{#2}}
\newcommand{\dlt}[1]{$#1$}
\newcommand{\bestdlt}[1]{\cellcolor{besthl}$\mathbf{#1}$}

\usepackage{graphicx}

\definecolor{twaccent}{HTML}{1F4E79}   
\newtcolorbox{takeaway2}[1][]{
  enhanced, breakable,
  colback=blue!4!white,
  frame hidden,
  borderline west={3.5pt}{0pt}{twaccent},
  left=10pt, right=8pt, top=7pt, bottom=7pt,
  before upper={%
    {\small\bfseries\scshape\textcolor{twaccent}{Key Takeaway}}%
    \par\vspace{3pt}%
    {\color{twaccent!40}\hrule height 0.4pt}%
    \vspace{5pt}%
  },
  fontupper=\normalsize,
  #1
}

\title{Structure Tax: How Structured Output affects LLMs Performance}
\author{Vineet Kumar \quad  Kanishka\thanks{Work done while at PayPal.} \quad Bhuvanesh Mandora \\
        PayPal Artificial Intelligence \\ PayPal, Bengaluru India \\
\texttt{vkumar32@paypal.com, kanishka21dhar@gmail.com, bmandora@paypal.com}
}

\begin{document}
\maketitle

\begin{abstract}
Deploying large language models in production often requires constraining outputs to structured formats such as JSON or XML, and prior work treats the resulting accuracy loss as an inherent `structure tax'.
We re-examine this claim by evaluating a battery of models, datasets and schemas, measuring task accuracy, confidence calibration, and hidden-state geometry.
The tax turns out to depend on schema design rather than on structure per se: reasoning-first field ordering matches or exceeds free-form accuracy, while answer-first ordering causes steep drops, particularly in smaller models.
Format sensitivity scales inversely with a task's own structural constraints, and schemas that preserve reasoning order also improve calibration with CKA showing greater separability between correct and incorrect representations in middle transformer layers.
Our findings indicate that properly designed structured formats can 
match or exceed free-form performance, reframing the critical question from `whether to structure' to `how to structure' for optimal reasoning preservation.
\end{abstract}

\section{Introduction}

Large language models  are now deployed across diverse domains, including mathematical reasoning, knowledge retrieval \citep{rathore2026cogni}, and code generation \cite{kumar2024genicious}. In practice, their utility depends not only on producing correct answers, but also on how outputs are structured and how reliably models communicate uncertainty. As a result, structured output formats such as XML tags, JSON schemas etc. have become common prompting strategies for improving reliability and downstream usability. However, their impact on model behavior remains underexplored. Do structured formats merely alter surface form, or do they change how models internally process and represent problems? And if structure affects reasoning, does it also influence confidence calibration?

LLMs are known to be sensitive to a wide range of prompt features separator tokens, casing, template choice, few-shot ordering and this brittleness has been documented across model scales and instruction-tuning regimes~\cite{sclar2024quantifying,he2024doespromptformattingimpact}. Within this broader landscape, output format constraints are distinctive on two counts. First, they are the part of a prompt that practitioners actually ship to production: free-form responses are rarely viable in deployed systems, so the output schema is not an optional choice but a structural commitment with downstream consequences. Second, unlike surface-token brittleness whose magnitude is empirical and whose direction is not predictable a priori the effect of output structure on reasoning admits a causal hypothesis grounded in the architecture itself. We therefore focus on field ordering as a mechanistically tractable sub-axis of format sensitivity, and ask whether the apparent cost of structured outputs reflects a property of structure per se or of a specific design choice within it.

In this work, we investigate \textbf{Structure Tax}: the systematic effect of output format constraints on task accuracy, calibration, and internal representations. Across five models and four benchmarks, we show that the apparent cost is not inherent to structure but contingent on a single design choice -- field ordering with second-order consequences for calibration and hidden-state geometry.

\begin{figure*}[hbt]
    \centering
    \includegraphics[width=0.98\textwidth]{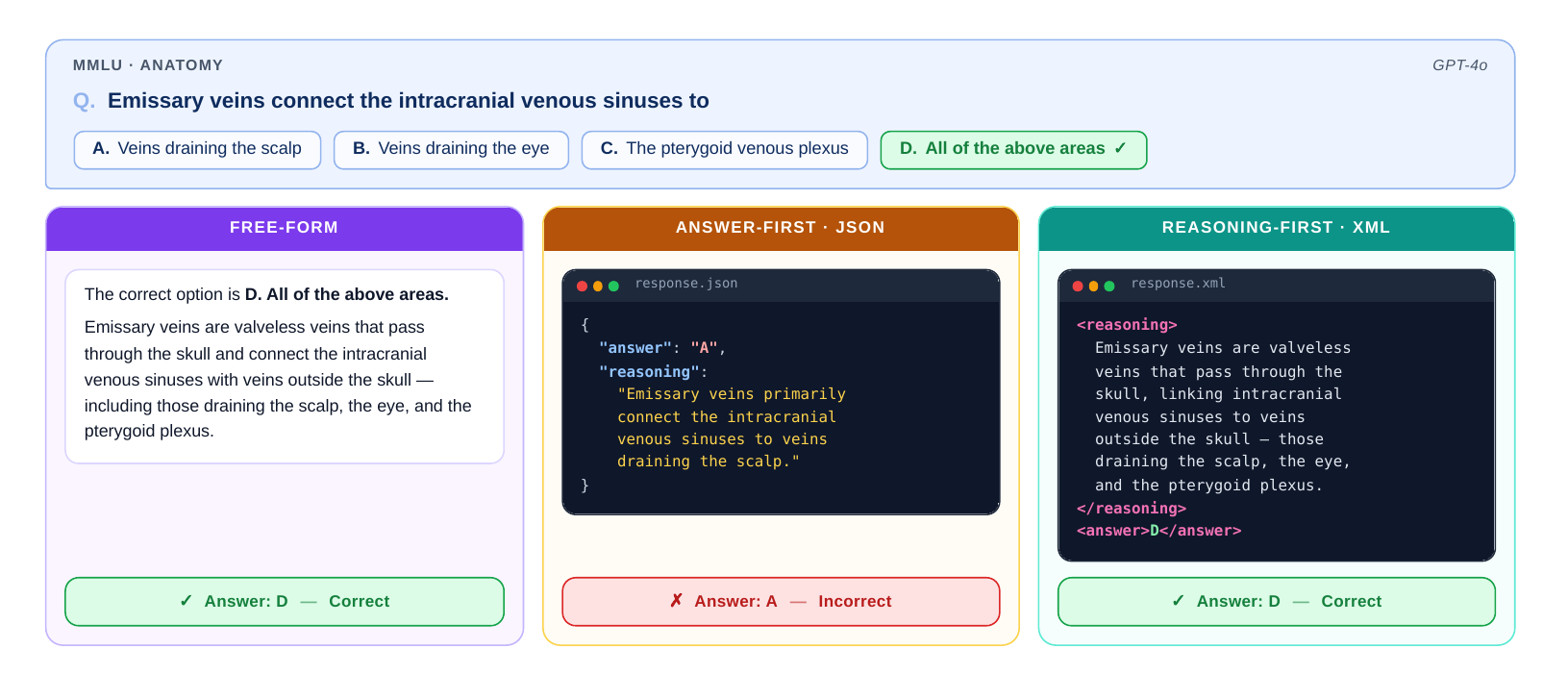}
  \caption{Three response formats elicited from GPT-4o on the same
  MMLU Anatomy question. \textbf{(Left)}~\emph{Free-form}: natural-language
  reasoning arrives at the correct answer \textbf{D}.
  \textbf{(Middle)}~\emph{Answer-first JSON}: the model commits to
  \texttt{"answer":\,"A"} before generating a post-hoc rationale that
  cites only one of three valid targets, yielding an incorrect answer.
  \textbf{(Right)}~\emph{Reasoning-first XML}: rationale unrolled inside
  \texttt{<reasoning>} precedes the commitment, recovering the correct
  \texttt{<answer>D</answer>}.}
    \label{fig:placeholder}
\end{figure*}

A key design dimension we examine is \emph{field ordering}: whether the reasoning chain (R) is placed before or after the final answer (A) within a structured schema. This distinction matters because autoregressive generation is strictly causal; each token conditions on all preceding tokens, but not on those yet to be produced. A \textit{reasoning-first} schema (R,A) forces the model to produce its reasoning trace before committing to an answer, allowing the answer token to be informed by a fully elaborated chain of thought. An \textit{answer-first} schema (A,R) inverts this dependency: the model commits to a specific answer token before any reasoning is generated, after which the reasoning field can only rationalize a decision already made. Figure~\ref{fig:placeholder} illustrates both orderings alongside a free-form baseline, making the structural difference concrete.

If field ordering matters because of autoregressive causality, then (R,A) schemas should preserve or improve reasoning quality while (A,R) schemas should degrade it, with the magnitude of degradation scaling with how much a task depends on multi-step inference. We confirm this prediction empirically. (R,A) schemas match or exceed free-form accuracy across the large majority of configurations, while (A,R) causes severe degradation on multi-step tasks, with the magnitude scaling inversely with a task's inherent structural organisation: low-structure tasks (knowledge QA, logical reasoning) benefit most; high-structure tasks (text-to-SQL) are largely format-invariant. This inverse-structure gradient holds at three levels of analysis --- task accuracy, confidence calibration, and hidden-state geometry via CKA and linear probes and structured formatting simultaneously reduces expected calibration error, establishing field ordering as both an accuracy and a calibration lever.

\section{Related Work}
\label{related_works}

\citet{tam-etal-2024-speak} report that JSON and XML output constraints can degrade reasoning accuracy, motivating the notion of a behavioural ``structure tax''; \citet{long-etal-2025-llms} broaden this to fifteen formats across four categories and propose mitigation through format-aware prompting and fine-tuning. A parallel body of work documents input-side sensitivity: \citet{sclar2024quantifying} show that semantically irrelevant surface features like separators, casing, spacing can swing accuracy with brittleness persisting across model scales and instruction tuning, and \citet{he2024doespromptformattingimpact} report a 40-point spread on GPT-3.5 from template variation alone. These studies treat each format as atomic asking \emph{which} format to use and lack a causal mechanism for the observed effects. We isolate a within-format axis, the relative position of reasoning and answer fields, where the direction of the effect is predictable from autoregressive generation: a token emitted at position $t$ can only condition on positions strictly less than $t$, so an answer token committed before its rationale cannot be revised by it.

The benefits of placing reasoning before the answer are established for chain-of-thought prompting~\cite{wei2022chain}. Most closely related, \citet{Fu_2026} compare answer-direct and reasoning-then-answer prompts on multiple-choice tasks and find that reasoning-first increases verbalised confidence regardless of correctness. Our work differs in three ways: we span a structural-density continuum rather than MCQ alone, we separate the format-family axis (JSON vs XML) from the field-ordering axis, and we trace the mechanism into hidden-state geometry. In a complementary fine-tuning setting, \citet{chen2025distillingreasoningabilitylarge} argue that post-thinking distillation (answer before rationale) is more robust to rationale errors for small models whether our prompted-generation finding persists under such training is an open question. Our results also sharpen the connection to chain-of-thought faithfulness~\cite{lanham2023measuring}: plausible-sounding reasoning can still be post-hoc if the answer token precedes it in the generation order.

Calibration is commonly assessed using Expected Calibration Error~\cite{minderer2021revisiting} and reliability diagrams~\cite{guo2017calibration}. Hidden states are known to encode correctness signals beyond surface outputs~\cite{servedio-etal-2025-hidden,orgad2024llms,zhang2025reasoning}, motivating combined behavioural-representational analysis. We compare layer representations using Centred Kernel Alignment~\cite{kornblith2019similarity}, mindful of documented interpretive caveats~\cite{davari2022reliability,cui2022deconfounded}, and verify CKA directions against linear probes and accuracy. To our knowledge, no prior work jointly examines how output format constraints affect task performance, calibration, and correctness-related representational geometry; our study addresses this gap.

\section{Experimental Design}
\label{expt_design}

We investigate three complementary aspects of format constraints   task accuracy, calibration, and internal representations   detailed in §\ref{sec:protocol}.


\subsection{Dataset}
\label{dataset}

We select four datasets that sample a continuum of intrinsic task structure the degree to which the task itself imposes compositional or sequential constraints on a valid answer, independent of any output format we impose. This continuum framing is central to our analysis: if format scaffolding interacts with intrinsic structure, the magnitude of any format effect should scale with where a task sits on the continuum. \textbf{Spider}~\cite{spider} anchors the high-structure end, a large-scale text-to-SQL dataset where SQL grammar imposes rigid compositional constraints on valid outputs. \textbf{GSM8K}~\cite{gsm8k} occupies a middle position: grade-school arithmetic word problems require multi-step reasoning that provides sequential scaffolding but no syntactic enforcement. \textbf{MMLU}~\cite{mmlu}, a 57-subject multitask benchmark, and \textbf{LogicBench}~\cite{logicbench}, designed to evaluate logical reasoning across reasoning types and complexity levels, sit at the low-structure end: the multiple-choice format is externally imposed and the reasoning path is otherwise unconstrained. JSON and XML are the two formats we evaluate; together they cover the principal axis a practitioner faces key-ordered serialisation (JSON) versus tag-bounded markup (XML) and span the formats most commonly shipped in production pipelines.

\subsection{Experiment protocol}
\label{sec:protocol}

Our experiment is three-pronged. We first measure task \textbf{accuracy under structured output} formats (JSON and XML, each in answer-first and reasoning-first orderings) against a free-form baseline. We then assess \textbf{calibration} by analyzing both verbalized confidence scores and token-level probability signals to determine whether format constraints shift the alignment between model confidence and empirical accuracy. Finally, we probe layer-wise hidden states using Centered Kernel Alignment (CKA) to characterize how format constraints alter the \textbf{internal geometry} of correct versus incorrect representations across model layers.

\paragraph{Why include the answer-first (A,R) ordering?}
We include (A,R) for three reasons. First, it mirrors a common production pattern where developers place the answer before the rationale because it reads more naturally to human reviewers (e.g., \textit{``rate this text and provide a rationale''}). Second, while the direction of the effect is intuitive, its \emph{magnitude} and \emph{model dependence} reqquires empirical characterization. Third, (A,R) isolates the effect of field ordering from the broader format-family distinction (JSON vs.\ XML), allowing us to attribute performance differences specifically to \emph{when} the answer is committed rather than to surface syntax.

\subsection{Models}
\label{models}

We evaluate five LLMs spanning both proprietary and open-weight families. The frontier proprietary models: \textbf{GPT-4o}~\cite{gpt4o}, \textbf{Gemini-Flash-2.0}~\cite{gemini-flash2}, and \textbf{Claude-Sonnet-4.5}~\cite{claude-sonnet45} -- span the three major frontier providers, reducing single-provider bias in our conclusions. The open-weight models -- \textbf{Llama-3.1-8B}~\cite{llama31} and \textbf{Gemma3-4B}~\cite{gemmateam2025gemma3technicalreport} span two scales (4B and 8B parameters) and two architecture families, which lets us check whether the representational patterns we observe (in Section §\ref{sec:cka}) generalise across architectures rather than depending on a single model design. The open-weight subset also enables reproducible, large-scale ablations under practical compute constraints.

\subsection{Evaluation Metrics}
Let $\hat{y}_i$ denote the model prediction and $p_i$ the confidence for input $x_i$ with ground truth $y_i$.

  \begin{itemize}
      \item \textbf{Accuracy} measures exact-match fraction; for Spider~\cite{spider} we use \textbf{Execution Accuracy (EX)}, which checks whether executing the predicted SQL yields the same result as the reference query.
      \item \textbf{ECE} quantifies calibration by comparing confidence to empirical accuracy across $M$ bins: \[\text{ECE} = \sum_{m=1}^{M} \frac{|B_m|}{N}\left|\text{acc}(B_m) - \text{conf}(B_m)\right|\]
      \item \textbf{Confidence--Correctness Correlation} ($\rho$) is the Spearman rank correlation between $p_i$ and $\mathbf{1}[\hat{y}_i = y_i]$, capturing whether higher confidence predicts correctness.
      \item Brier Score and Entropy, used in Appendix~\ref{sec:app_elicitation} and~\ref{sec:app_logit}, are defined there.
  \end{itemize}

\section{Structure Tax}
\label{main_results}

\begin{table*}[h!]
\centering
\small
\setlength{\tabcolsep}{5pt}
\renewcommand{\arraystretch}{1.25}
\begin{tabular}{@{}ll ccc cc cc @{\hspace{2pt}}}

\toprule
& & & \multicolumn{2}{c}{\textbf{Answer-first (A,R)}} & \multicolumn{2}{c}{\textbf{Reasoning-first (R,A)}} \\
\cmidrule(lr){4-5} \cmidrule(lr){6-7}
\textbf{Dataset} & \textbf{Model} & \textbf{Free Form} & \textbf{JSON} & \textbf{XML} & \textbf{JSON} & \textbf{XML} \\
\midrule
\multirow{5}{*}{\textbf{GSM8K}}
    & GPT-4o            & \cell{88.78}{1.02} & \cell{55.34}{1.58} & \cell{56.86}{2.71} & \cell{95.60}{1.01} & \best{95.60}{0.17} \\
    & Gemini-Flash-2.0  & \cell{79.38}{0.28} & \cell{63.08}{0.53} & \cell{54.90}{3.64} & \best{95.53}{0.59} & \cell{95.45}{0.60} \\
    & Claude-Sonnet-4.5 & \cell{86.50}{1.13} & \cell{96.13}{0.88} & \cell{62.35}{2.29} & \cell{87.79}{0.98} & \best{97.65}{0.39} \\
    & {Llama-3.1-8B}      & \cell{79.98}{1.10} & \cell{31.01}{1.58} & \cell{10.98}{2.66} & \cell{80.21}{0.45} & \best{83.53}{2.58} \\
    & {Gemma3-4B}         & \cell{78.82}{1.46} & \cell{27.45}{4.70} & \cell{10.98}{2.13} & \cell{83.53}{1.96} & \best{87.06}{2.34} \\
\addlinespace[2pt]
\midrule
\multirow{5}{*}{\textbf{LogicBench}}
    & GPT-4o            & \cell{79.00}{0.96} & \cell{83.80}{1.29} & \cell{83.00}{1.29} & \cell{83.20}{0.73} & \best{84.20}{1.93} \\
    & Gemini-Flash-2.0  & \cell{72.80}{2.11} & \cell{76.80}{1.64} & \cell{84.77}{2.34} & \cell{82.40}{1.75} & \best{85.40}{1.16} \\
    & Claude-Sonnet-4.5 & \cell{87.60}{2.17} & \cell{56.60}{2.70} & \best{88.40}{0.55} & \cell{75.00}{0.77} & \cell{87.20}{1.35} \\
    & {Llama-3.1-8B}      & \cell{57.20}{0.79} & \best{76.00}{1.05} & \cell{70.34}{1.36} & \cell{74.60}{2.05} & \cell{73.00}{2.50} \\
    & {Gemma3-4B}         & \cell{61.20}{0.71} & \best{69.40}{1.33} & \cell{67.40}{1.87} & \cell{66.00}{1.61} & \cell{66.20}{2.22} \\
\addlinespace[2pt]
\midrule
\multirow{5}{*}{\textbf{MMLU}}
    & GPT-4o            & \cell{79.95}{0.84} & \cell{82.50}{0.83} & \cell{81.58}{1.36} & \best{87.46}{0.70} & \cell{86.94}{0.78} \\
    & Gemini-Flash-2.0  & \cell{56.24}{0.78} & \cell{84.52}{0.61} & \cell{83.96}{0.94} & \cell{84.26}{1.13} & \best{86.14}{0.76} \\
    & Claude-Sonnet-4.5 & \cell{85.37}{0.98} & \cell{90.84}{0.65} & \cell{90.66}{0.63} & \cell{90.92}{0.32} & \best{92.03}{0.64} \\
    & {Llama-3.1-8B}      & \cell{62.83}{0.80} & \cell{64.14}{1.71} & \cell{55.98}{0.75} & \best{71.26}{0.70} & \cell{68.06}{0.86} \\
    & {Gemma3-4B}         & \cell{59.76}{1.08} & \cell{60.29}{0.83} & \cell{58.72}{1.08} & \best{63.10}{1.03} & \cell{62.12}{1.54} \\
\addlinespace[2pt]
\midrule
\multirow{5}{*}{\textbf{Spider}}
    & GPT-4o            & \cell{69.15}{1.78} & \cell{69.05}{1.32} & \cell{71.76}{1.57} & \cell{71.47}{0.93} & \best{73.02}{1.40} \\
    & Gemini-Flash-2.0  & \best{79.40}{0.84} & \cell{77.37}{1.33} & \cell{78.20}{1.09} & \cell{76.69}{1.34} & \cell{77.18}{1.33} \\
    & Claude-Sonnet-4.5 & \cell{77.47}{1.60} & \cell{77.47}{1.23} & \best{78.05}{1.20} & \cell{77.37}{1.36} & \cell{77.66}{1.52} \\
    & {Llama-3.1-8B}      & \best{64.99}{1.85} & \cell{58.03}{0.86} & \cell{64.31}{1.15} & \cell{54.84}{2.71} & \cell{62.19}{1.60} \\
    & {Gemma3-4B}         & \best{70.21}{1.35} & \cell{66.15}{1.48} & \cell{68.38}{0.89} & \cell{67.50}{0.97} & \cell{63.35}{0.57} \\
\bottomrule
\end{tabular}
\caption{Impact of output structure constraints on accuracy across benchmark tasks. We compare \textit{Answer-first} (answer before reasoning) and \textit{Reasoning-first} (reasoning before answer) orderings under JSON and XML formats, against an unconstrained Free Form baseline. Values are mean accuracy over 5 independent runs (standard deviation in gray). Best format per model--dataset pair is \textbf{bolded} and highlighted.}
\label{tab:results}
\end{table*}

\label{structure-tax}

We ask whether imposing a structured output schema incurs a universal accuracy penalty   a \textit{structure tax}   or whether its effect depends on how the schema is designed.

\paragraph{Setup.}
Each model receives a task-specific system prompt specifying the required output schema (JSON or XML; R,A or A,R); all other prompt content is held constant across conditions. Responses are collected at \texttt{temperature\,=\,0} via each provider's API. Evaluation uses exact-match accuracy for GSM8K, MMLU, and LogicBench, and execution accuracy (EX) for Spider. Prompt templates are reproduced in Appendix~\ref{sec:xml-json-prompts}.

\paragraph{Results (Table~\ref{tab:results}).}
Reasoning-first formats dominate. XML\,(R,A) or JSON\,(R,A) wins the best-format cell in 13 of 20 model-dataset pairs and never collapses on tasks where A,R wins. The A,R penalty is sharpest on GSM8K XML\,(A,R) falls to 10.98\% for both \textbf{\textbf{Llama-3.1-8B}} and \textbf{\textbf{Gemma3-4B}}, against 83-87\% under (R,A); confirming that committing the answer token before the reasoning chain breaks multi-step arithmetic entirely. On Spider (the high-structure task), freeform remains competitive for three of five models, consistent with SQL generation providing its own intrinsic structure.

\paragraph{Analysis.}
The severity of the ordering effect scales with task reasoning depth: the R,A advantage is largest on GSM8K, moderate on MMLU and LogicBench, and negligible on Spider. Within (R,A)  XML outperforms JSON (margin up to 12.2\,pp on LogicBench/Claude-Sonnet-4.5), suggesting that explicit tag boundaries scaffold the reasoning-answer transition more reliably than JSON key ordering alone. Smaller models are more fragile: {{Llama-3.1-8B}} and {{Gemma3-4B}} show the largest collapses under A,R and higher run-to-run variance, whereas frontier models are sensitive to ordering but do not catastrophically fail. Crucially, R,A formats simultaneously improve accuracy {and} reduce variance. 

\paragraph{Robustness to surface prompt variation.} A natural concern is whether the (R,A) advantage is an artefact of the specific prompts used. We test this by perturbing instructional phrasing, schema casing, and whitespace across three variants per condition on GSM8K with Llama-3.1-8B and GPT-4o (full setup in Appendix~\ref{sec:app_brittleness}). The (R,A) > (A,R) gap is preserved in every (model, format-family, variant) cell: the smallest gap across all 24 combinations is +37 points (GPT-4o JSON), while within-cell spread under perturbation is at most 11.5 points. Field ordering's effect on accuracy is qualitatively invariant to surface prompt changes, distinguishing it from the format-template brittleness documented by \cite{sclar2024quantifying} and \cite{he2024doespromptformattingimpact}.

\begin{takeaway2}
Field ordering, not structure itself, governs the structure tax: reasoning-first schemas (R,A) consistently match or nearly match the best-performing formats, while answer-first schemas (A,R) may cause severe and in some settings catastrophic degradation on reasoning-heavy tasks.
\end{takeaway2}

\section{Calibration and Confidence Estimation}

\label{sec:calibration}

  Accuracy gains from structured formatting are only meaningful if models also communicate reliable uncertainty.
  We ask whether output format affects calibration i.e. alignment between stated confidence and empirical accuracy.
  Confidence is elicited as a verbal 1--10 integer score and calibration is measured via ECE; elicitation strategy and logit-verbal alignment are explained in Appendix~\ref{sec:app_calib}. Because the accuracy experiments in Section~\ref{main_results} identify XML with reasoning-first ordering, (R,A), as the recommended structured format, we evaluate the calibration of this configuration against free-form generation. The results show that the recommended XML (R,A) format also reduces expected calibration error relative to free-form output.

  \subsection{Output Format as a Calibration Mechanism}
  \label{sec:calib_format}

\begin{figure}[h!]
      \centering
      \includegraphics[width=\linewidth]{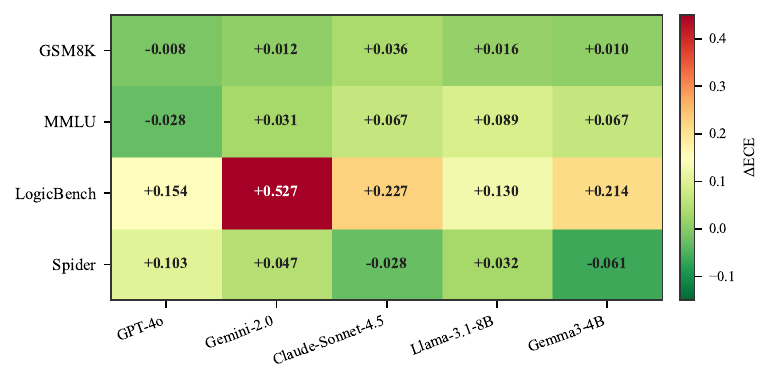}
      \caption{ECE increase under freeform relative to XML ($\Delta\text{ECE} = \text{ECE}_\text{freeform} -
  \text{ECE}_\text{XML}$). Positive values (red) indicate freeform is worse calibrated; negative (green) indicates
  the reverse. LogicBench shows the largest format sensitivity; GSM8K is format-invariant.}
      \label{fig:ece_delta}
  \end{figure}

  Figure~\ref{fig:ece_delta} reports $\Delta\text{ECE} = \text{ECE}_\text{freeform} - \text{ECE}_\text{XML}$; positive values indicate freeform is \emph{worse} calibrated.
  The pattern closely mirrors the accuracy results: format sensitivity is highest on tasks lacking inherent structure and negligible where intrinsic scaffolding already exists.
  LogicBench shows consistent gains under XML across all models; MMLU shows gains for four of five models, the exception being GPT-4o where freeform is marginally better calibrated. GSM8K is effectively format-invariant — arithmetic provides its own sequential discipline that format cannot augment.  
  Spider (the high-structure task) is the dataset where freeform is better calibrated for some model, again consistent with SQL imposing its own structural constraints.
  These results establish XML formatting not merely as an accuracy intervention but as a \emph{calibration mechanism}: it regularises confidence expression alongside reasoning quality.

\vspace{-0.1in}
  \subsection{Calibration Profiles Under Structured Output}
  \label{sec:calib_xml}

  Table~\ref{tab:calibration} reports ECE, overconfidence rate (OvConf), and Spearman $\rho$ between confidence and correctness under XML prompting.
  The task-level ordering is stable across models: GSM8K is best calibrated, Spider is worst, with LogicBench and MMLU in between.
  This ordering reflects how much intrinsic task structure guides confidence. Arithmetic answers are either right or wrong in a verifiable chain, while SQL execution correctness is opaque to the model at generation time
  Across all settings, the confidence-accuracy correlation $\rho$ remains low, indicating that while XML improves aggregate calibration, per-example discrimination is limited.  Appendix Figures~\ref{fig:fig_conf_acc_scatter} and~\ref{fig:fig_reliability_diagrams} visualize the confidence--accuracy gap and per-bin calibration across all model--dataset conditions.

\begin{table}[h!]
\centering
\small
\setlength{\tabcolsep}{6pt}
\renewcommand{\arraystretch}{1.25}
\resizebox{0.475\textwidth}{!}{%
\begin{tabular}{@{}ll ccc}
\toprule
\textbf{Dataset} & \textbf{Model} & \textbf{ECE}\,$\downarrow$ & \textbf{OvConf}\,$\downarrow$ & \textbf{Corr}\,($\rho$)\,$\uparrow$ \\
\midrule
\multirow{5}{*}{\textbf{GSM8K}}
    & GPT-4o             & \cell{0.050}{0.005} & \cell{0.051}{0.005} & \cell{0.169}{0.028} \\
    & Gemini-Flash-2.0   & \cell{0.042}{0.002} & \cell{0.043}{0.002} & \cell{0.134}{0.084} \\
    & Claude-Sonnet-4.5  & \best{0.020}{0.004} & \best{0.022}{0.004} & \cell{0.264}{0.063} \\
    & Llama-3.1-8B       & \cell{0.136}{0.008} & \cell{0.138}{0.009} & \cell{0.189}{0.053} \\
    & Gemma3-4B          & \cell{0.109}{0.028} & \cell{0.111}{0.029} & \best{0.350}{0.146} \\
\addlinespace[2pt]
\midrule
\multirow{5}{*}{\textbf{LogicBench}}
    & GPT-4o             & \cell{0.097}{0.018} & \cell{0.197}{0.018} & \cell{0.253}{0.078} \\
    & Gemini-Flash-2.0   & \cell{0.097}{0.017} & \cell{0.131}{0.022} & \best{0.291}{0.049} \\
    & Claude-Sonnet-4.5  & \best{0.060}{0.009} & \best{0.108}{0.012} & \cell{0.251}{0.037} \\
    & Llama-3.1-8B       & \cell{0.103}{0.016} & \cell{0.244}{0.020} & \cell{0.222}{0.064} \\
    & Gemma3-4B          & \cell{0.243}{0.015} & \cell{0.337}{0.015} & \cell{0.256}{0.042} \\
\addlinespace[2pt]
\midrule
\multirow{5}{*}{\textbf{MMLU}}
    & GPT-4o             & \cell{0.073}{0.005} & \cell{0.132}{0.006} & \cell{0.289}{0.023} \\
    & Gemini-Flash-2.0   & \cell{0.104}{0.010} & \cell{0.131}{0.011} & \cell{0.282}{0.040} \\
    & Claude-Sonnet-4.5  & \best{0.041}{0.007} & \best{0.082}{0.004} & \best{0.396}{0.026} \\
    & Llama-3.1-8B       & \cell{0.208}{0.014} & \cell{0.321}{0.012} & \cell{0.173}{0.021} \\
    & Gemma3-4B          & \cell{0.292}{0.016} & \cell{0.365}{0.014} & \cell{0.214}{0.026} \\
\addlinespace[2pt]
\midrule
\multirow{5}{*}{\textbf{Spider}}
    & GPT-4o             & \cell{0.201}{0.018} & \cell{0.258}{0.017} & \cell{0.247}{0.025} \\
    & Gemini-Flash-2.0   & \cell{0.210}{0.011} & \best{0.211}{0.010} & \cell{0.115}{0.032} \\
    & Claude-Sonnet-4.5  & \best{0.185}{0.016} & \cell{0.235}{0.015} & \cell{0.299}{0.038} \\
    & Llama-3.1-8B       & \cell{0.273}{0.009} & \cell{0.351}{0.009} & \cell{0.171}{0.018} \\
    & Gemma3-4B          & \cell{0.283}{0.006} & \cell{0.326}{0.008} & \best{0.331}{0.050} \\
\bottomrule
\end{tabular}
}
\caption{Calibration under XML-structured output. ECE and OvConf are lower-is-better; Corr ($\rho$) is higher-is-better. Values are mean over 5 runs (std.\ in gray). Best model per metric--dataset pair is \textbf{bolded}.}
\label{tab:calibration}
\end{table}

  \paragraph{Elicitation strategy and logit-verbal alignment.}
Absolute elicitation generally outperforms distributional prompting and is used as the default (Appendix~\ref{sec:app_elicitation}). Verbal scores correlate moderately with logits only in stronger models; smaller models show little alignment, with large KL divergence between verbal and token-level uncertainty signals (Appendix~\ref{sec:app_logit}). Thus, for smaller models, verbalized confidence is better viewed as a task-conditioned heuristic than a reflection of internal uncertainty.

\begin{takeaway2}
Output format acts as a calibration mechanism, not just an accuracy intervention: XML (R,A) reduces ECE alongside improving reasoning quality, with gains following the same inverse-structure gradient observed in Section~\ref{main_results}. 
\end{takeaway2}

\section{Internal Representations and Mechanism}
  \label{sec:cka}

The accuracy and calibration results establish \emph{what} field ordering does; this section asks \emph{why}.
We open the model's intermediate computations with two complementary tools.
\textbf{CKA} \cite{kornblith2019similarity} acts as a representational separability thermometer: given hidden states for correct and incorrect predictions at each layer, lower cross-group CKA means the model has built more geometrically distinct internal representations for the two outcomes.
\textbf{Linear probes} act as an information readout: a linear classifier trained on hidden states at layer $\ell$ tells us whether the answer identity is already decodable at that depth, before it is generated.
A third lens   reasoning length   connects the surface output to the representational evidence.
All three converge on the same account of how field ordering shapes computation.

  \subsection{Representational Separability: CKA}
  \label{sec:calib_cka}

  We apply CKA to \textbf{Llama-3.1-8B} \& \textbf{Gemma3-4B} under XML\,(R,A), XML\,(A,R), and freeform across all four datasets; extraction details are in Appendix~\ref{sec:app_cka}.

\begin{figure*}[h!]
    \centering
    \begin{subfigure}[t]{0.48\linewidth}
        \centering
        \includegraphics[width=\linewidth]{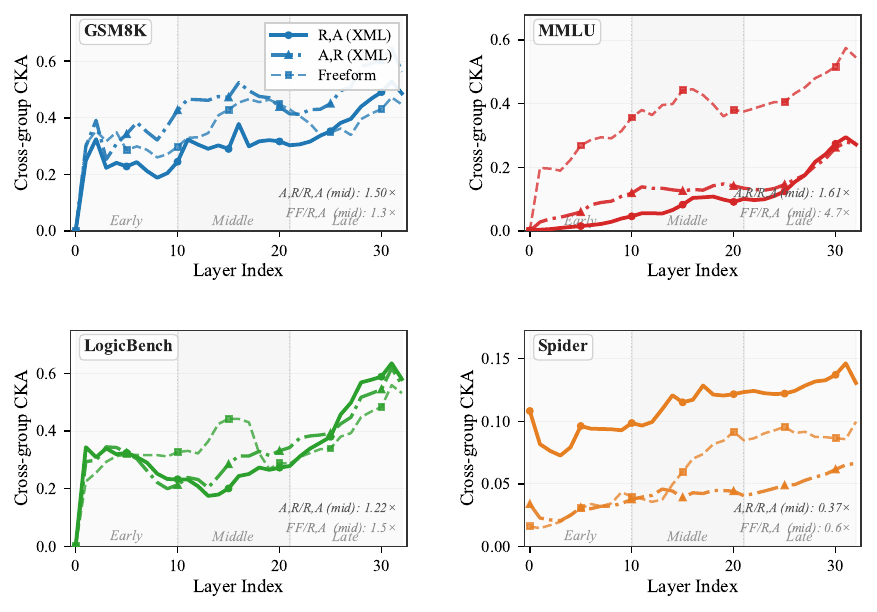}
        \caption{Cross-group CKA per layer}
        \label{fig:cka_main}
    \end{subfigure}
    \hfill
    \begin{subfigure}[t]{0.48\linewidth}
        \centering
        \includegraphics[width=\linewidth]{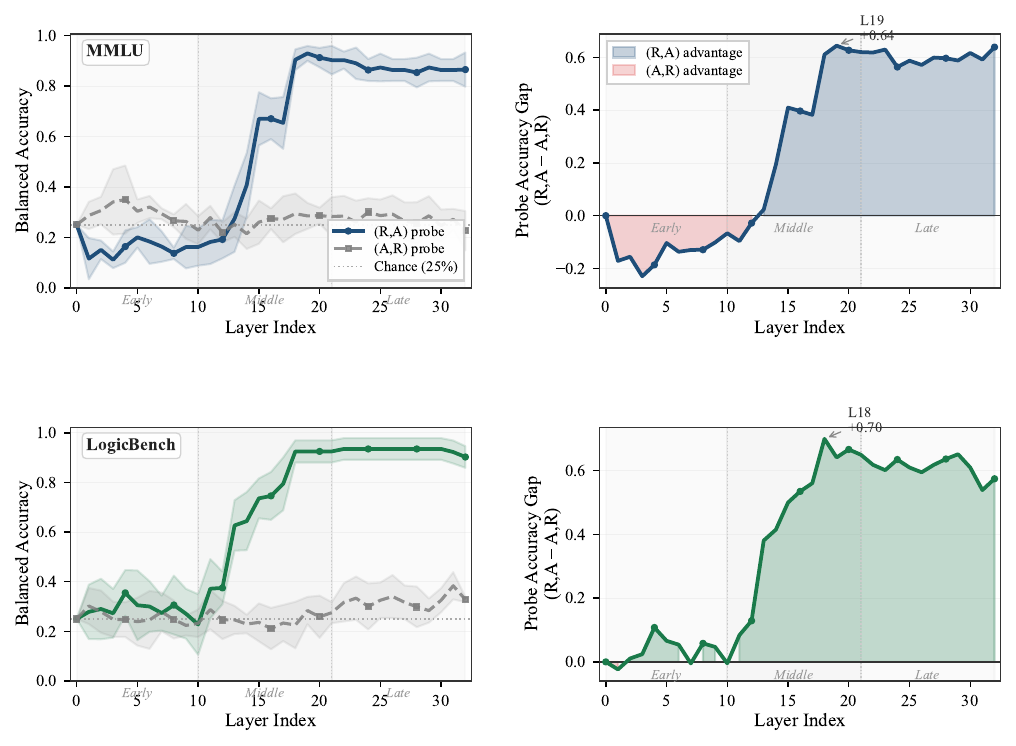}
        \caption{Linear probe balanced accuracy per layer}
        \label{fig:probe_main}
    \end{subfigure}
    \caption{Internal representations for \textbf{Llama-3.1-8B} on MMLU and LogicBench.
    \textbf{Left}: Cross-group CKA under XML\,(R,A) (solid), XML\,(A,R) (dash-dot), and Freeform (dashed) across four benchmarks. Higher CKA = weaker separation between correct and incorrect representations; R,A yields the lowest mid-layer CKA on reasoning tasks while Spider reverses the pattern.
    \textbf{Right}: Linear probe balanced accuracy decoding the predicted answer label from hidden states at each layer. Solid = (R,A); dashed = (A,R); dotted = chance; gap panel shows the (R,A)$-$(A,R) advantage with peak layer annotated.
    \textbf{Gemma3-4B} replicates both patterns at reduced magnitude (Appendix~\ref{sec:app_gemma_representations}).}
    \label{fig:representations}
\end{figure*}

  Figure~\ref{fig:representations} (left) shows that format sensitivity tracks the same inverse-structure gradient seen in accuracy and calibration.
  On \textbf{MMLU}   the task with the least inherent structure   freeform cross-group CKA rises steadily from early layers and stays elevated, whereas XML\,(R,A) keeps it below $0.15$ through the middle layers: the model is building representations that separate right from wrong only when the output schema forces deliberate reasoning first.
  \textbf{LogicBench} shows a moderate version of the same pattern, converging in late layers as both conditions reach the same reasoning conclusion.
  \textbf{GSM8K} exhibits near-parity across formats   arithmetic provides its own sequential scaffold   while \textbf{Spider} reverses the gradient entirely, the only task where freeform yields lower cross-group CKA than XML, consistent with SQL imposing its own compositional structure that XML tags disrupt.
  Critically, the direction of the CKA effect correctly predicts the direction of the accuracy effect in all four tasks: Spider is the only task where XML reduces accuracy and the only task where the CKA ratio falls below parity.
  \textbf{Gemma3-4B} replicates this pattern at reduced magnitude (Appendix~\ref{sec:app_gemma_representations}), confirming findings are not model-specific.

An instructive dissociation appears on MMLU: despite the largest representational suppression under freeform prompting, Llama gains only modest accuracy compared to LogicBench. This highlights the limit of format scaffolding i.e., accuracy remains constrained by the model’s underlying knowledge. XML can enforce deliberate ordering, not supply missing facts. Within-group CKA heatmaps further show that XML induces clearer stage-wise block structure in correct representations (Appendix, Figures~\ref{fig:cka_heatmaps_gsm8k_mmlu}--\ref{fig:cka_heatmaps_logic_spider}).

  \subsection{Answer Encoding: Linear Probes}
  \label{sec:probes}

  While CKA measures whether representations are separable, probes ask a more targeted question: does the model already \emph{know the answer} during the reasoning phase, before the answer token is generated?
  We train a linear classifier on pre-answer hidden states at each layer to decode the predicted label and compare accuracy under (R,A) vs (A,R) on MMLU and LogicBench (GSM8K and Spider excluded; see Appendix~\ref{sec:app_probes}).

  The answer is unambiguous (Figure~\ref{fig:representations}, right).
  Under (R,A), answer identity becomes linearly decodable from middle layers onward, peaking at balanced accuracy of $0.64$ (MMLU, L19) and $0.70$ (LogicBench, L18)   well above chance and sustained across the final third of the network.
  Under (A,R), the same probe stagnates near chance through early and middle layers: the model has committed to an answer token but has not built a representation of \emph{which} answer it chose during the reasoning phase   the answer is present in the output stream. With the answer field first, the model must emit the answer letter without preceding explicit reasoning. The pre-answer-letter state may therefore reflect less task-relevant computation, increasing the risk that commitment precedes deliberate inference. 
  \textbf{Gemma3-4B} replicates the directional pattern at reduced magnitude (Appendix~\ref{sec:app_gemma_representations}), again confirming this is a general property of field ordering, not an artefact of one model family.

  \subsection{Reasoning Length and the Two Failure Modes}
  \label{sec:length}

  The probe and CKA results describe what happens inside the model; reasoning length reveals the surface signature.
  We measure word count across 80 cells (5 models $\times$ 4 datasets $\times$ 4 format-orderings) as a proxy for reasoning effort; extraction details and full results appear in Appendix~\ref{sec:app_length}.

  The data reveal two qualitatively distinct failure modes of the (A,R) ordering.
  The first is \textbf{rationalization}: on MCQ tasks, committing the answer token causes the reasoning field to collapse   JSON\,(A,R) produces just 33 words on MMLU against 121 under XML\,(R,A), a fourfold compression into a brief post-hoc justification rather than deliberate inference.
  The second is \textbf{premature commitment}: on GSM8K, reasoning length is flat across all four orderings (108--127 words), yet accuracy collapses from $0.92$ under R,A to $0.55$ under JSON\,(A,R)   the model writes as much, but arithmetic that should determine the answer is instead forced to justify one already committed.

  Length alone does not predict accuracy ($r=+0.04$ overall); the ordering dimension carries the signal.
  JSON\,(A,R)'s correlation of $r=-0.73$ ($p<0.001$)   the only significant result across all four format-orderings (Appendix Table~\ref{tab:length_corr})   arises because the cells where JSON\,(A,R) does \emph{not} compress (GSM8K, Spider) are precisely those where accuracy is worst: verbose post-commitment reasoning is worse than compressed post-commitment reasoning, because ordering determines whether effort precedes or follows the decision.

  \begin{takeaway2}
CKA, linear probes, and reasoning-length analysis all support the same conclusion: reasoning-first ordering lets the model form clear answer-related representations before committing to a final response, preserving deliberate reasoning. In contrast, answer-first ordering leads to either compressed post-hoc rationalization in MCQ tasks or incorrect computation despite long reasoning traces in arithmetic tasks. XML\,(R,A) avoids both by allowing the answer to be generated only after the reasoning process is fully developed.
\end{takeaway2}

\section{Discussion and Conclusion}
\label{conclusion}
\paragraph{Two failure modes: practical diagnostics.}
The two failure modes share a single underlying cause autoregressive generation forces the answer token to be sampled either before or after the reasoning trace but they call for different diagnostics in deployment.
Rationalization on MCQ tasks is detectable from output length alone: a sharp drop in reasoning-field token count under (A,R) relative to (R,A) is a reliable signal that the model is producing post-hoc justification rather than deliberate inference.
Premature commitment on arithmetic tasks is harder to detect, because reasoning length is preserved (108--127 words across all four orderings on GSM8K); here the diagnostic must check whether the reasoning is consistent with the stated answer, not merely whether reasoning is present.
Practitioners auditing structured pipelines for reasoning failures should therefore monitor both the volume of the reasoning field and its semantic alignment with the committed answer.

The ``structure tax'' attributed to structured output constraints is largely an artefact of answer-first field ordering rather than structure per se. Autoregressive generation is strictly causal, so an answer token emitted before its rationale cannot be revised by it; reasoning-first (R,A) schemas avoid this and consistently match or exceed free-form accuracy across five models and four datasets, while answer-first schemas produce either compressed post-hoc rationalization on MCQ tasks or premature commitment on arithmetic tasks. The format-sensitivity gradient is consistent across accuracy (Table~\ref{tab:results}), calibration (Figure~\ref{fig:ece_delta}), and hidden-state geometry (Figure~\ref{fig:representations}), with gains largest on low-structure tasks and negligible on high-structure tasks.

\paragraph{Scope and applicability.}
The gains we report concentrate where the task lacks intrinsic structure and the model is mid-sized: low-structure benchmarks (MMLU, LogicBench) under smaller open-weight models show the largest absolute (R,A) advantages and the most dramatic (A,R) collapses. On high-structure tasks like text-to-SQL, where compositional constraints are imposed by the output language itself, field ordering matters less and the choice between (R,A) and free-form is close to a wash. Frontier models are sensitive to ordering but rarely catastrophic under (A,R), suggesting larger models partially compensate for the causal-ordering constraint. For practitioners, reasoning-first XML is therefore the safer default: it never meaningfully hurts on the tasks where structure matters, and it preserves accuracy and calibration on the tasks where it does---without sacrificing downstream parseability.

\section*{Limitations}

A recurring pattern across our analyses is that format sensitivity scales inversely with a task's inherent structural organisation Spider at the high-structure end, MMLU and LogicBench at the low-structure end, GSM8K between them.
Our four datasets sample this continuum at only four coarse points, which is sufficient to establish the inverse-scaling trend but insufficient to formally characterise a \emph{sweet spot} --- the level of intrinsic structural density at which format scaffolding is maximally beneficial.
Formally locating such a sweet spot would require systematic variation of structural density within a single task family (e.g., varying constraint density in SQL generation, or step-count and carry-chain length in arithmetic), holding all other factors constant.
We flag this as a concrete and important direction for future work.

Representational analysis via CKA and linear probes is restricted to two open-weight models ({Llama-3.1-8B} and {Gemma3-4B}), leaving open whether the observed middle-layer patterns generalise across model scales and architectures.
We examine only two field orderings (answer-first and reasoning-first); real-world schemas involve finer-grained structure such as planning fields, intermediate steps, and verification stages, and how structural granularity beyond binary ordering affects reasoning remains an open question.
All experiments rely on prompting rather than training-time adaptation, so it is unclear whether fine-tuning on structured formats would produce the same representational geometry.
Finally, extending the analysis cross-lingually may reveal whether format scaffolding compensates more strongly in settings where the model has weaker internalised task structure.


\bibliography{custom}

\appendix

\label{sec:appendix}

\clearpage

\onecolumn

{%
  \section*{Appendix Contents}
  \vspace{-0.4em}\hrule height 0.4pt\vspace{0.6em}
  \small
  \newcommand{\tocsec}[2]{%
    \noindent\textbf{\hyperref[#1]{#2}}%
    \nobreak\hfill\textbf{\pageref{#1}}\par\vspace{0.15em}}
  \newcommand{\tocsub}[2]{%
    \noindent\hspace{1.4em}\hyperref[#1]{#2}%
    \nobreak\hfill\pageref{#1}\par\vspace{0.05em}}

  \tocsec{sec:prompts}{A\quad Complete Prompt Collection}
    \tocsub{sec:xml-json-prompts}{A.1\quad Structured Format Evaluation}
    \tocsub{sec:calibration-prompts}{A.2\quad Calibration Analysis Prompts}
    \tocsub{sec:Schema-violation}{A.3\quad Schema violation statistics}
    \tocsub{sec:logit-prompts}{A.4\quad Logit Confidence Analysis Prompts}
  \tocsec{sec:app_datasets}{B\quad Dataset Summary Statistics}
  \tocsec{sec:app_calib}{C\quad Calibration: Supplementary Details}
    \tocsub{sec:app_calib_impl}{C.1\quad Implementation Details}
    \tocsub{sec:app_elicitation}{C.2\quad Confidence Elicitation: Absolute vs.\ Relative}
    \tocsub{sec:app_logit}{C.3\quad Verbal Confidence vs.\ Internal Probabilities}
    \tocsub{sec:app_calib_viz}{C.4\quad Calibration Visualisations}
  \tocsec{sec:app_representations}{D\quad CKA and Probe: Method Details}
    \tocsub{sec:app_cka}{D.1\quad CKA Extraction}
    \tocsub{sec:app_cka_heatmaps}{D.2\quad Within-Group CKA Heatmaps: Layout and Reading Guide}
    \tocsub{sec:app_probes}{D.3\quad Linear Probe Setup}
    \tocsub{sec:app_gemma_representations}{D.4\quad Gemma3-4B: Representational Results}
    \tocsec{sec:app_length}{E\quad Reasoning Length: Measurement and Results}
    \tocsec{sec:app_brittleness}{F\quad Prompt-Brittleness Ablation}
  \vspace{0.3em}\hrule height 0.4pt
}
\vspace{0.8em}

\vspace{0.6em}
\section{Appendix: Complete Prompt Collection}
\label{sec:prompts}

This appendix documents all prompts used in our three experimental
conditions: Structured Format Evaluation~(\S\ref{sec:xml-json-prompts}),
Calibration Analysis~(\S\ref{sec:calibration-prompts}), and Logit
Confidence Analysis~(\S\ref{sec:logit-prompts}).


\subsection{Structured Format Evaluation}
\label{sec:xml-json-prompts}

\paragraph{GSM8K (Mathematical Reasoning).}

\begin{mdframed}[style=xmlbox]
\small\ttfamily
\textbf{[XML AR]} You are a math expert. Solve step-by-step in XML format.\\[2pt]
\hspace*{1em}\textless{}answer\textgreater{} \textit{final numerical answer} \textless{}/answer\textgreater{}\\
\hspace*{1em}\textless{}reasoning\textgreater{} \textit{step-by-step solution} \textless{}/reasoning\textgreater{}\\[2pt]
Problem: \{problem\}
\end{mdframed}

\begin{mdframed}[style=xmlbox]
\small\ttfamily
\textbf{[XML RA]} You are a math expert. Solve step-by-step in XML format.\\[2pt]
\hspace*{1em}\textless{}reasoning\textgreater{} \textit{step-by-step solution} \textless{}/reasoning\textgreater{}\\
\hspace*{1em}\textless{}answer\textgreater{} \textit{final numerical answer} \textless{}/answer\textgreater{}\\[2pt]
Problem: \{problem\}
\end{mdframed}

\begin{mdframed}[style=jsonbox]
\small\ttfamily
\textbf{[JSON AR]} You are a math expert. Solve step-by-step in JSON format.\\[2pt]
\hspace*{1em}\{ "answer": "final answer", "reasoning": "solution" \}\\[2pt]
Problem: \{problem\}
\end{mdframed}

\begin{mdframed}[style=loosebox]
\small\ttfamily
\textbf{[JSON RA]} You are a math expert. Solve step-by-step in JSON format.\\[2pt]
\hspace*{1em}\{ "reasoning": "solution", "answer": "final answer" \}\\[2pt]
Problem: \{problem\}
\end{mdframed}

\begin{mdframed}[style=freebox]
\small\ttfamily
\textbf{[Freeform]} You are a math expert. Solve step-by-step.\\[2pt]
Problem: \{problem\}
\end{mdframed}

\paragraph{MMLU (Multiple-Choice Knowledge).}

\begin{mdframed}[style=xmlbox]
\small\ttfamily
\textbf{[XML AR]} You are a helpful AI assistant. Analyze and provide your
response in XML format.\\[2pt]
\hspace*{1em}\textless{}answer\textgreater{} \textit{option letter (A--D)} \textless{}/answer\textgreater{}\\
\hspace*{1em}\textless{}reasoning\textgreater{} \textit{analysis} \textless{}/reasoning\textgreater{}\\[2pt]
Problem: \{problem\}\quad Options: \{options\}
\end{mdframed}

\begin{mdframed}[style=xmlbox]
\small\ttfamily
\textbf{[XML RA]} You are a helpful AI assistant. Analyze and provide your
response in XML format.\\[2pt]
\hspace*{1em}\textless{}reasoning\textgreater{} \textit{analysis and thought process} \textless{}/reasoning\textgreater{}\\
\hspace*{1em}\textless{}answer\textgreater{} \textit{option letter (A--D)} \textless{}/answer\textgreater{}\\[2pt]
Problem: \{problem\}\quad Options: \{options\}
\end{mdframed}

\begin{mdframed}[style=jsonbox]
\small\ttfamily
\textbf{[JSON AR]} You are a helpful AI assistant. Analyze and provide your
response in JSON format.\\[2pt]
\hspace*{1em}\{ "answer": "option letter (A--D)", "reasoning": "explanation" \}\\[2pt]
Problem: \{problem\}\quad Options: \{options\}
\end{mdframed}

\begin{mdframed}[style=loosebox]
\small\ttfamily
\textbf{[JSON RA]} You are a helpful AI assistant. Analyze and provide your
response in JSON format.\\[2pt]
\hspace*{1em}\{ "reasoning": "explanation", "answer": "option letter (A--D)" \}\\[2pt]
Problem: \{problem\}\quad Options: \{options\}
\end{mdframed}

\begin{mdframed}[style=freebox]
\small\ttfamily
\textbf{[Freeform]} You are a helpful AI assistant. Solve this MCQ.\\[2pt]
Problem: \{problem\}\quad Options: \{options\}
\end{mdframed}

\paragraph{LogicBench (Logical Reasoning).}

\begin{mdframed}[style=xmlbox]
\small\ttfamily
\textbf{[XML AR]} You are a reasoning expert. Answer in XML format.\\[2pt]
\hspace*{1em}\textless{}answer\textgreater{} \textit{option letter} \textless{}/answer\textgreater{}\\
\hspace*{1em}\textless{}reasoning\textgreater{} \textit{analysis} \textless{}/reasoning\textgreater{}\\[2pt]
Context: \{context\}\quad Problem: \{question\}\quad Options: \{options\}
\end{mdframed}

\begin{mdframed}[style=xmlbox]
\small\ttfamily
\textbf{[XML RA]} You are a reasoning expert. Answer in XML format.\\[2pt]
\hspace*{1em}\textless{}reasoning\textgreater{} \textit{analysis} \textless{}/reasoning\textgreater{}\\
\hspace*{1em}\textless{}answer\textgreater{} \textit{option letter} \textless{}/answer\textgreater{}\\[2pt]
Context: \{context\}\quad Problem: \{question\}\quad Options: \{options\}
\end{mdframed}

\begin{mdframed}[style=jsonbox]
\small\ttfamily
\textbf{[JSON AR]} You are a reasoning expert. Answer in JSON format.\\[2pt]
\hspace*{1em}\{ "answer": "option letter", "reasoning": "explanation" \}\\[2pt]
Context: \{context\}\quad Problem: \{question\}\quad Options: \{options\}
\end{mdframed}

\begin{mdframed}[style=loosebox]
\small\ttfamily
\textbf{[JSON RA]} You are a reasoning expert. Answer in JSON format.\\[2pt]
\hspace*{1em}\{ "reasoning": "explanation", "answer": "option letter" \}\\[2pt]
Context: \{context\}\quad Problem: \{question\}\quad Options: \{options\}
\end{mdframed}

\begin{mdframed}[style=freebox]
\small\ttfamily
\textbf{[Freeform]} You are a reasoning expert. Answer the following question.\\[2pt]
Context: \{context\}\quad Problem: \{question\}\quad Options: \{options\}
\end{mdframed}

\paragraph{Spider (Text-to-SQL).}

\begin{mdframed}[style=xmlbox]
\small\ttfamily
\textbf{[XML AR]} You are a SQL expert. Generate query in XML format.\\[2pt]
\hspace*{1em}\textless{}answer\textgreater{} \textit{SQL query} \textless{}/answer\textgreater{}\\
\hspace*{1em}\textless{}reasoning\textgreater{} \textit{rationale} \textless{}/reasoning\textgreater{}\\[2pt]
Schema: \{schema\}\quad Question: \{question\}
\end{mdframed}

\begin{mdframed}[style=xmlbox]
\small\ttfamily
\textbf{[XML RA]} You are a SQL expert. Generate query in XML format.\\[2pt]
\hspace*{1em}\textless{}reasoning\textgreater{} \textit{rationale} \textless{}/reasoning\textgreater{}\\
\hspace*{1em}\textless{}answer\textgreater{} \textit{SQL query} \textless{}/answer\textgreater{}\\[2pt]
Schema: \{schema\}\quad Question: \{question\}
\end{mdframed}

\begin{mdframed}[style=jsonbox]
\small\ttfamily
\textbf{[JSON AR]} You are a SQL expert. Return query in JSON.\\[2pt]
\hspace*{1em}\{ "answer": "SQL query", "reasoning": "explanation" \}\\[2pt]
Schema: \{schema\}\quad Question: \{question\}
\end{mdframed}

\begin{mdframed}[style=loosebox]
\small\ttfamily
\textbf{[JSON RA]} You are a SQL expert. Return query in JSON.\\[2pt]
\hspace*{1em}\{ "reasoning": "explanation", "answer": "SQL query" \}\\[2pt]
Schema: \{schema\}\quad Question: \{question\}
\end{mdframed}

\begin{mdframed}[style=freebox]
\small\ttfamily
\textbf{[Freeform]} You are a SQL expert. Generate a SQL query.\\[2pt]
Schema: \{schema\}\quad Question: \{question\}
\end{mdframed}

\subsection{Calibration Analysis Prompts}
\label{sec:calibration-prompts}

Calibration prompts extend each format with an explicit confidence
request on a 1--10 scale. Below we show the XML and Freeform variants;
JSON calibration follows the same JSON structure as \S\ref{sec:xml-json-prompts}
with an additional \texttt{"confidence"} field.

\paragraph{XML Calibration.}
All four datasets share the same XML structure extended by a
\texttt{<confidence>} field:

\begin{mdframed}[style=xmlbox]
\small\ttfamily
\textbf{[XML + Confidence]} \textit{(role and task instruction as above)}\\[2pt]
\hspace*{1em}\textless{}reasoning\textgreater{} \textit{analysis} \textless{}/reasoning\textgreater{}\\
\hspace*{1em}\textless{}answer\textgreater{} \textit{answer} \textless{}/answer\textgreater{}\\
\hspace*{1em}\textless{}confidence\textgreater{} \textit{integer 1--10 (1\,=\,not confident,}
\textit{10\,=\,very confident)} \textless{}/confidence\textgreater{}\\[2pt]
\textit{Question fields: identical to Structured Format prompts above.}
\end{mdframed}

\paragraph{Freeform Calibration.}

\begin{mdframed}[style=freebox]
\small\ttfamily
\textbf{[Freeform + Confidence --- MMLU]}\\
You are a helpful AI assistant. Answer the question directly. Before you
finish, state your confidence level on a scale of 1--10 (1\,=\,not
confident, 10\,=\,very confident).\\[2pt]
Problem: \{problem\}\quad Options: \{options\}
\end{mdframed}

\begin{mdframed}[style=freebox]
\small\ttfamily
\textbf{[Freeform + Confidence --- GSM8K]}\\
You are a math expert. Solve step-by-step. At the end, state your
confidence level on a scale of 1--10.\\[2pt]
Problem: \{problem\}
\end{mdframed}

\begin{mdframed}[style=freebox]
\small\ttfamily
\textbf{[Freeform + Confidence --- LogicBench]}\\
You are a reasoning expert. Answer the following question. State your
confidence on a scale of 1--10.\\[2pt]
Context: \{context\}\quad Problem: \{question\}\quad Options: \{options\}
\end{mdframed}

\begin{mdframed}[style=freebox]
\small\ttfamily
\textbf{[Freeform + Confidence --- Spider]}\\
You are a SQL expert. Generate a SQL query. Before you finish, state
your confidence on a scale of 1--10.\\[2pt]
Schema: \{schema\}\quad Question: \{question\}
\end{mdframed}

\subsection{Schema-violation statistics}
\label{sec:Schema-violation}
Structured-output schema violations were zero for all experiments. The only parsing failures occurred in free-form outputs due to the Gemma3-4B.
\begin{table}[h!]
\centering
\small
\caption{Free-form empty-extraction rates caused by the Gemma3-4B parser bug.}
\label{tab:empty_extraction}
\begin{tabular}{lc}
\toprule
\textbf{Dataset} & \textbf{Empty Extraction Rate} \\
\midrule
MMLU        & \textbf{29.8\%} \\
LogicBench  & \textbf{45.2\%} \\
\bottomrule
\end{tabular}
\end{table}
\subsection{Logit Confidence Analysis Prompts}
\label{sec:logit-prompts}

Logit-based confidence extraction uses minimal prompts that elicit a
single-token answer, enabling direct extraction of confidence from the
token probability distribution over \{A, B, C, D\}.

\begin{mdframed}[style=logitbox]
\small\ttfamily
\textbf{[Logit --- MMLU \& LogicBench]}\\
Answer the following question with only the single letter of the correct
option (A, B, C, or D). Do not include any explanation or punctuation
--- just the letter.\\[2pt]
Problem: \{problem\}\quad Options: \{options\}\quad Answer:
\end{mdframed}

\paragraph{Confidence extraction.}
For logit prompts, we record the softmax probability assigned to each
option token at the \texttt{Answer:} position. For XML/JSON prompts, we
parse the \texttt{<confidence>} or \texttt{"confidence"} field and
normalise to $[0,1]$. For freeform prompts, confidence is extracted via
regex (e.g., \texttt{"confidence[:\textbackslash{}s]+(\textbackslash{}d+)"}).

\medskip
\noindent\textbf{Reproducibility.} All experiments use
temperature\,$=$\,0.0 to ensure deterministic outputs. Prompts are
identical across models; no model-specific tuning was applied.
\section{Dataset summary statistics}
\label{sec:app_datasets}  
\begin{table}[h]
  \centering
  \small
  \renewcommand{\arraystretch}{1.2}
  \begin{tabular}{llccc}
  \toprule
  \textbf{Dataset} & \textbf{Split} & \textbf{Examples} & \textbf{Task Type} & \textbf{Answer Format} \\
  \midrule
  GSM8K       & Test       & 1{,}319 & Math reasoning     & Numeric           \\
  LogicBench  & Eval       & 500     & Logical inference  & MCQ        \\
  MMLU        & Validation & 1{,}531 & Knowledge QA       & MCQ        \\
  Spider      & Validation & 1{,}030 & Text-to-SQL        & SQL query         \\
  \bottomrule
  \end{tabular}
  \caption{Dataset statistics and properties. All splits are held-out evaluation sets; no training data is used.
  Spider is evaluated via execution accuracy; all other datasets use exact-match accuracy.}
  \label{tab:datasets}
  \end{table}


\section{Calibration: Supplementary Details}
\label{sec:app_calib}

\subsection{Implementation Details}
\label{sec:app_calib_impl}
Calibration is measured with Expected Calibration Error (ECE) using $M=10$ equal-width bins over $[0,1]$, computed as
$\text{ECE} = \sum_{m=1}^{10}\frac{|B_m|}{N}|\text{acc}(B_m)-\text{conf}(B_m)|$,
where the final bin is closed on both sides.
Confidence is elicited as a verbal integer on a 1--10 scale (``1 = not confident, 10 = very confident'') and normalized by dividing by 10.
Scores are extracted from free-text responses via regex matching on the pattern \texttt{confidence[:\textbackslash s]+(\textbackslash d+)}.
All API calls use \texttt{temperature\,=\,0}.
\textbf{Brier Score} measures mean squared error between predicted probabilities and binary correctness: $\frac{1}{N}\sum_i(p_i - y_i)^2$.
\textbf{Entropy} $H(p) = -\sum_y p(y|x_i)\log p(y|x_i)$ measures predictive uncertainty; used under relative confidence elicitation.

\subsection{Confidence Elicitation: Absolute vs.\ Relative}
\label{sec:app_elicitation}

We compare two elicitation strategies on MMLU and LogicBench (MCQ tasks only, as relative prompting requires a fixed option set).
\textbf{Absolute} prompting requests a single scalar score.
\textbf{Relative} prompting asks the model to distribute confidence across the answer options (A/B/C/D), from which we extract the chosen-option probability.
Table~\ref{tab:elicitation} reports ECE, average confidence, entropy, and Brier score.

Absolute prompting achieves lower or equal ECE in 6 of 8 model--dataset pairs.
Relative prompting produces lower average confidence ($-0.035$ to $-0.16$) and higher entropy ($0.47$--$1.14$), indicating models express greater distributional uncertainty, but this does not translate into better-calibrated scalar estimates for most models.
The Brier score under relative prompting is consistently worse for weaker models (\textbf{Llama-3.1-8B}: $0.452$--$0.582$), while Claude-Sonnet-4.5 achieves the lowest Brier score under relative prompting on both datasets, suggesting stronger models can exploit distributional elicitation more effectively.
\textbf{Gemma3-4B} is an exception to the absolute-wins pattern: relative prompting reduces its ECE by $\Delta = +0.138$ (LogicBench) and $+0.083$ (MMLU), the two largest per-model improvements in the comparison, consistent with its tendency toward overconfident scalar responses.

\begin{table*}[h!]
\centering
\small
\setlength{\tabcolsep}{4pt}
\renewcommand{\arraystretch}{1.25}
\begin{tabular}{@{}ll ccc cc cc}
\toprule
& & \multicolumn{3}{c}{\textbf{ECE}\,$\downarrow$} & \multicolumn{2}{c}{\textbf{AvgConf}} & \multirow{2}{*}{\textbf{Entropy}\,$\downarrow$} & \multirow{2}{*}{\textbf{Brier}\,$\downarrow$} \\
\cmidrule(lr){3-5} \cmidrule(lr){6-7}
\textbf{Dataset} & \textbf{Model} & Abs. & Rel. & $\Delta^{\dagger}$ & Abs. & Rel. & & \\
\midrule
\multirow{5}{*}{\textbf{LogicBench}}
    & GPT-4o            & \cell{0.072}{.010} & \cell{0.084}{.017} & \dlt{-0.012} & \cell{0.883}{.006} & \cell{0.741}{.005} & \cell{1.141}{.015} & \cell{0.297}{.023} \\
    & Gemini-Flash-2.0  & \cell{0.094}{.012} & \best{0.052}{.005} & \dlt{+0.042} & \cell{0.941}{.004} & \cell{0.798}{.004} & \cell{0.930}{.027} & \cell{0.244}{.014} \\
    & Claude-Sonnet-4.5 & \best{0.045}{.007} & \cell{0.066}{.010} & \dlt{-0.021} & \cell{0.879}{.007} & \cell{0.834}{.003} & \cell{0.759}{.014} & \best{0.200}{.016} \\
    & {Llama-3.1-8B}      & \cell{0.166}{.015} & \cell{0.161}{.015} & \dlt{+0.005} & \cell{0.914}{.006} & \cell{0.878}{.007} & \best{0.496}{.019} & \cell{0.452}{.036} \\
    & {Gemma3-4B}         & \cell{0.290}{.012} & \cell{0.152}{.017} & \bestdlt{+0.138} & \cell{0.938}{.002} & \cell{0.775}{.002} & \cell{1.046}{.010} & \cell{0.530}{.019} \\
\addlinespace[2pt]
\midrule
\multirow{5}{*}{\textbf{MMLU}}
    & GPT-4o            & \cell{0.063}{.007} & \cell{0.117}{.007} & \dlt{-0.053} & \cell{0.924}{.002} & \cell{0.750}{.005} & \cell{1.142}{.010} & \cell{0.244}{.011} \\
    & Gemini-Flash-2.0  & \cell{0.103}{.008} & \cell{0.078}{.011} & \dlt{+0.025} & \cell{0.969}{.002} & \cell{0.803}{.004} & \cell{0.921}{.010} & \cell{0.242}{.007} \\
    & Claude-Sonnet-4.5 & \best{0.024}{.005} & \best{0.057}{.005} & \dlt{-0.033} & \cell{0.933}{.003} & \cell{0.864}{.003} & \best{0.671}{.002} & \best{0.147}{.007} \\
    & {Llama-3.1-8B}      & \cell{0.247}{.008} & \cell{0.192}{.010} & \dlt{+0.055} & \cell{0.927}{.001} & \cell{0.834}{.004} & \cell{0.682}{.004} & \cell{0.582}{.018} \\
    & {Gemma3-4B}         & \cell{0.281}{.016} & \cell{0.198}{.023} & \bestdlt{+0.083} & \cell{0.948}{.002} & \cell{0.744}{.012} & \cell{1.093}{.023} & \cell{0.591}{.027} \\
\bottomrule
\end{tabular}
\caption{Absolute vs.\ relative confidence elicitation on MCQ datasets. \textsuperscript{$\dagger$}$\Delta = \text{ECE}_{\text{Abs}} - \text{ECE}_{\text{Rel}}$; positive = relative prompting improves calibration. Entropy and Brier defined only under relative prompting. Highlighted $\Delta$ values mark the largest per-column improvements. Values are mean over 5 runs (std.\ in gray).}
\label{tab:elicitation}
\end{table*}

\subsection{Verbal Confidence vs.\ Internal Probabilities}
\label{sec:app_logit}

We compare verbal 1--10 scores against logit-based confidence   the softmax probability assigned to the selected answer token   on MMLU and LogicBench for GPT-4o, \textbf{Llama-3.1-8B}, and \textbf{Gemma3-4B} (the models for which logits are accessible).

\begin{table*}[h!]
  \centering
  \small
  \begin{tabular}{llcccccccc}
    \toprule
    & & \multicolumn{3}{c}{\textbf{Correlation \& Divergence}} & \multicolumn{2}{c}{\textbf{ECE}~$\downarrow$} & \multicolumn{2}{c}{\textbf{Brier Score}~$\downarrow$} \\
    \cmidrule(lr){3-5}\cmidrule(lr){6-7}\cmidrule(lr){8-9}
    \textbf{Dataset} & \textbf{Model} & Spearman~$\rho$ & Pearson~$r$ & Mean KL & Logit & Verbal & Logit & Verbal \\
    \midrule
    \multirow{3}{*}{\textbf{MMLU}}
      & GPT-4o       & $0.439$ & $0.128$ & $0.500$ & $0.117$ & $0.116$ & $0.276$ & ${0.247}$ \\
      & {Llama-3.1-8B} & $-0.049^{\dagger}$ & $-0.040^{\dagger}$ & $3.617$ & ${0.156}$ & $0.191$ & ${0.496}$ & $0.582$ \\
      & {Gemma3-4B}    & $0.024^{\dagger}$  & $0.002^{\dagger}$  & $1.676$ & $0.402$ & $0.212$ & $0.813$ & $0.602$ \\
    \midrule
    \multirow{3}{*}{\textbf{LogicBench}}
      & GPT-4o       & $0.673$ & $0.312$ & $0.423$ & $0.122$ & ${0.084}$ & ${0.274}$ & $0.305$ \\
      & {Llama-3.1-8B} & $0.282$ & $0.230$ & $2.304$ & ${0.141}$ & $0.162$ & ${0.379}$ & $0.453$ \\
      & {Gemma3-4B}    & $0.148$ & $0.099$ & $0.757$ & $0.311$ & $0.146$ & $0.627$ & $0.532$ \\
    \bottomrule
  \end{tabular}
  \caption{Logit vs.\ verbal confidence comparison. Spearman $\rho$ and Pearson $r$ measure scalar agreement on the chosen-option score; Mean KL measures full distributional divergence. ECE and Brier assess calibration against correctness. $^\dagger$~$p > 0.05$.}
  \label{tab:logit_verbal}
\end{table*}

Results show a clear model-size gradient.
GPT-4o verbal and logit confidence are moderately correlated ($r_s = 0.44$ on MMLU, $0.67$ on LogicBench) with low KL divergence ($\approx 0.5$), suggesting verbal scores partially reflect internal probability estimates.
\textbf{Llama-3.1-8B} breaks down on MMLU ($r_s = -0.05$, $p = .07$, KL $= 3.62$): verbal confidence is effectively decoupled from token-level probabilities, yet logit ECE ($0.156$) is better than verbal ECE ($0.191$), making logits the more reliable signal for this model.
\textbf{Gemma3-4B} is the most extreme case: near-zero Spearman on MMLU ($r_s = 0.024$, not significant, KL $= 1.676$), and logit ECE ($0.402$) is far worse than verbal ECE ($0.212$)   logits are dominated by the token-level distribution rather than calibrated uncertainty, while verbal confidence, though imprecise, is at least weakly correlated with accuracy.
On LogicBench, all models improve: GPT-4o reaches $r_s = 0.67$, and even \textbf{Gemma3-4B} achieves a moderate Spearman ($0.148$), suggesting task difficulty moderates verbal-logit alignment.

\subsection{Calibration Visualisations}
\label{sec:app_calib_viz}

\begin{figure*}[h!]
    \centering
    \includegraphics[width=1\linewidth]{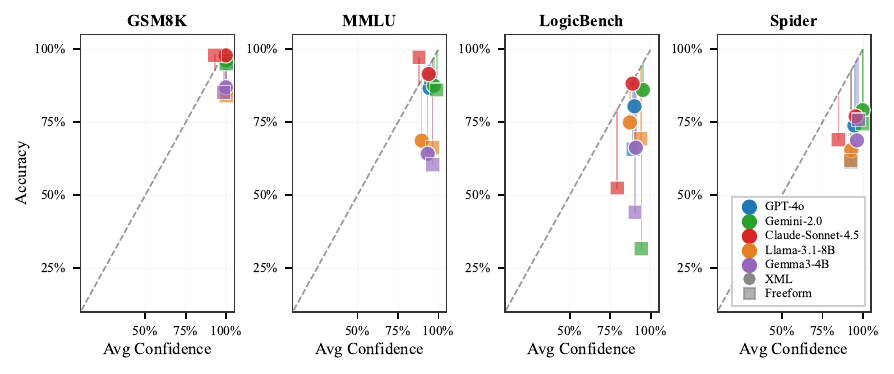}
    \caption{  Accuracy vs. confidence scatter for all 5 models across 4 datasets under freeform (squares) and XML
  (circles) prompting. All models express high confidence ($>85\%$) regardless of format, while accuracy  
  varies substantially. The gap between confidence and the diagonal is consistently larger under
  freeform, most dramatically for \textbf{Gemma3-4B} on MMLU and LogicBench ($<$26\% accuracy, $\sim$100\%          
  confidence) and \textbf{Llama-3.1-8B} on LogicBench (23\% accuracy, 93\% confidence). XML structured output
  reduces overconfidence across all models, bringing points closer to the calibration diagonal.}
    \label{fig:fig_conf_acc_scatter}
\end{figure*}

Figure~\ref{fig:fig_conf_acc_scatter} plots mean confidence against mean accuracy for each model--dataset combination under freeform (squares) and XML (circles).
All models cluster in the high-confidence region ($>85\%$) regardless of format while accuracy spans a wide range, placing most points well above the diagonal that marks perfect calibration.
The format effect appears as a systematic downward-and-rightward shift: XML moves points closer to the diagonal, reducing the confidence--accuracy gap across all conditions.
The displacement is largest for weaker models on low-structure tasks, consistent with the ECE reductions in Table~\ref{tab:calibration}.
GSM8K points cluster near the diagonal under both formats, confirming it as the best-calibrated task irrespective of schema.

\begin{figure*}[h!]
    \centering
    \includegraphics[width=1\linewidth]{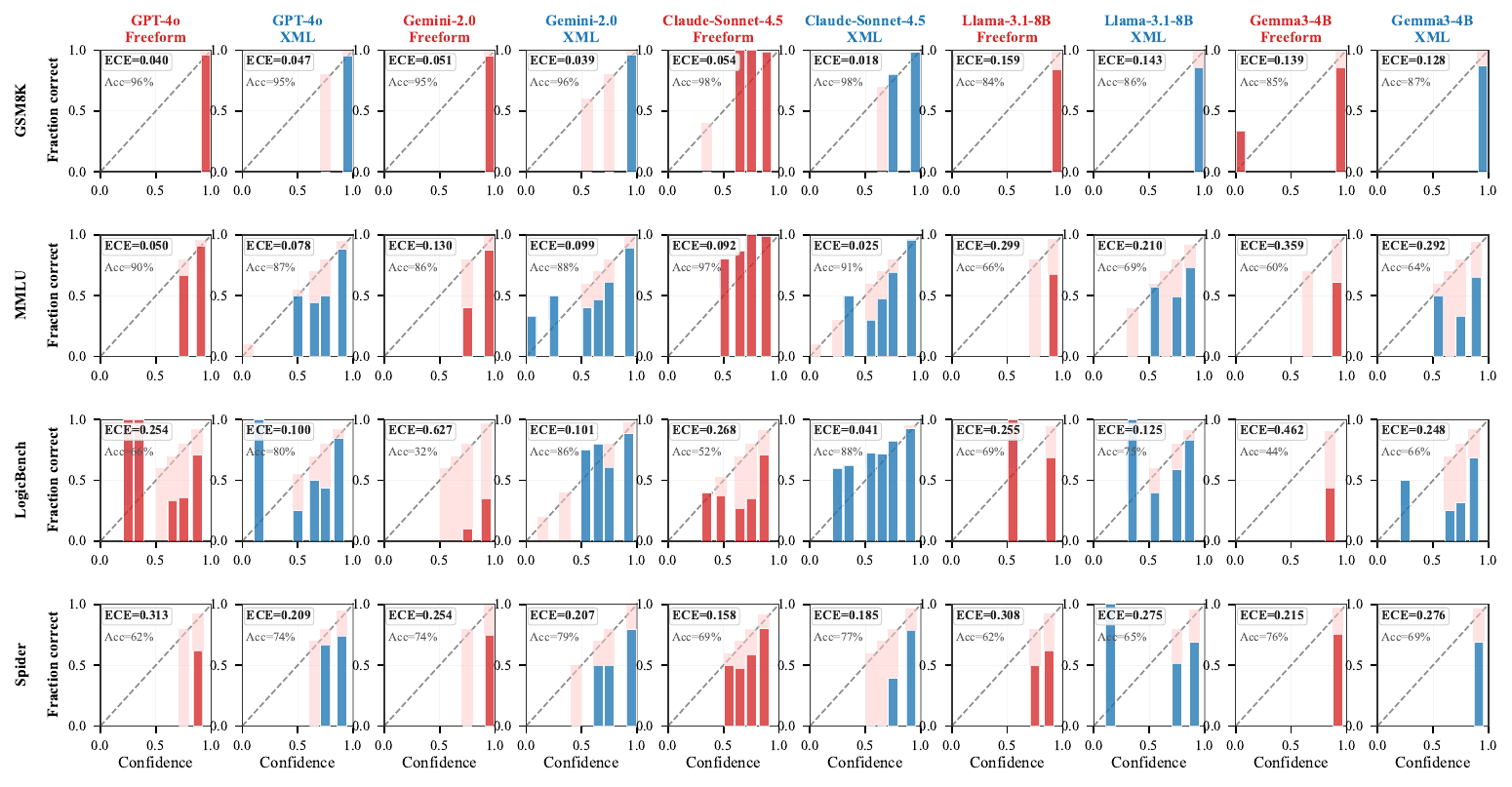}
    \caption{  Reliability diagrams comparing freeform and XML calibration across 20 model–dataset combinations (5    
  models × 4 datasets, each shown in freeform and XML format). XML bars align closer to the diagonal in  
  most configurations, with the largest improvements on LogicBench where freeform produces severe
  overconfidence (ECE up to 0.703 for \textbf{Llama-3.1-8B} and 0.645 for \textbf{Gemma3-4B}). \textbf{Gemma3-4B} shows the highest 
  overall freeform ECE, particularly on MMLU (0.721). GSM8K remains comparatively well-calibrated under
  both formats across all models.}
    \label{fig:fig_reliability_diagrams}
\end{figure*}

Figure~\ref{fig:fig_reliability_diagrams} shows per-bin reliability diagrams for all 20 model--dataset conditions.
Under freeform, high-confidence bins are heavily populated while empirical accuracy lags, producing bars far below the diagonal   the visual signature of overconfidence.
XML shifts bars toward the diagonal across most configurations, with the largest recalibration on LogicBench and MMLU where freeform calibration is worst.
GSM8K bars are near-diagonal under both formats, reinforcing the format-invariance of well-structured tasks.
Together, the two figures confirm that the ECE reductions in Table~\ref{tab:calibration} and Figure~\ref{fig:ece_delta} reflect genuine per-bin recalibration rather than an artefact of aggregation.

\newpage
\section{CKA and Probe: Method Details}
\label{sec:app_representations}

\subsection{CKA Extraction}
\label{sec:app_cka}

For each question we run a single forward pass through the frozen model with no generation and extract the \emph{last-token} hidden state at every layer $\ell$, yielding representation matrices $\mathbf{X}^\ell_{\mathrm{C}}, \mathbf{X}^\ell_{\mathrm{I}} \in \mathbb{R}^{n \times d}$ for the correct and incorrect prediction groups respectively ($n=200$ examples per dataset, sampled uniformly).
Inputs are formatted identically to the main benchmark runs i.e. XML or freeform, ensuring representational differences are attributable to output format, not input variation.
We apply this to Llama-3.1-8B ($d=4096$, 33 layers) and Gemma3-4B ($d=2560$, 35 layers).

We compute linear CKA via the HSIC estimator:
\[
\text{CKA}(\mathbf{X},\mathbf{Y}) = \frac{\text{HSIC}(\mathbf{X}\mathbf{X}^\top,\mathbf{Y}\mathbf{Y}^\top)}{\sqrt{\text{HSIC}(\mathbf{X}\mathbf{X}^\top,\mathbf{X}\mathbf{X}^\top)\cdot\text{HSIC}(\mathbf{Y}\mathbf{Y}^\top,\mathbf{Y}\mathbf{Y}^\top)}}
\]
where $\mathrm{HSIC}(K,L)=\frac{1}{(n-1)^2}\mathrm{tr}(K_c L_c)$ and subscript $c$ denotes centering.
This yields three quantities per layer: (i) a per-layer \emph{cross-group} CKA scalar measuring geometric alignment between correct and incorrect representations; (ii) a within-correct inter-layer CKA matrix; and (iii) an analogous within-incorrect matrix.
We partition layers into three processing stages Early, Middle, and Late at roughly equal thirds of network depth.

\begin{figure*}[p]  
      \centering
      \setlength{\abovecaptionskip}{4pt}
      \setlength{\belowcaptionskip}{0pt}

      \begin{subfigure}{\textwidth}
          \includegraphics[width=\textwidth, height=0.421\textheight, keepaspectratio]{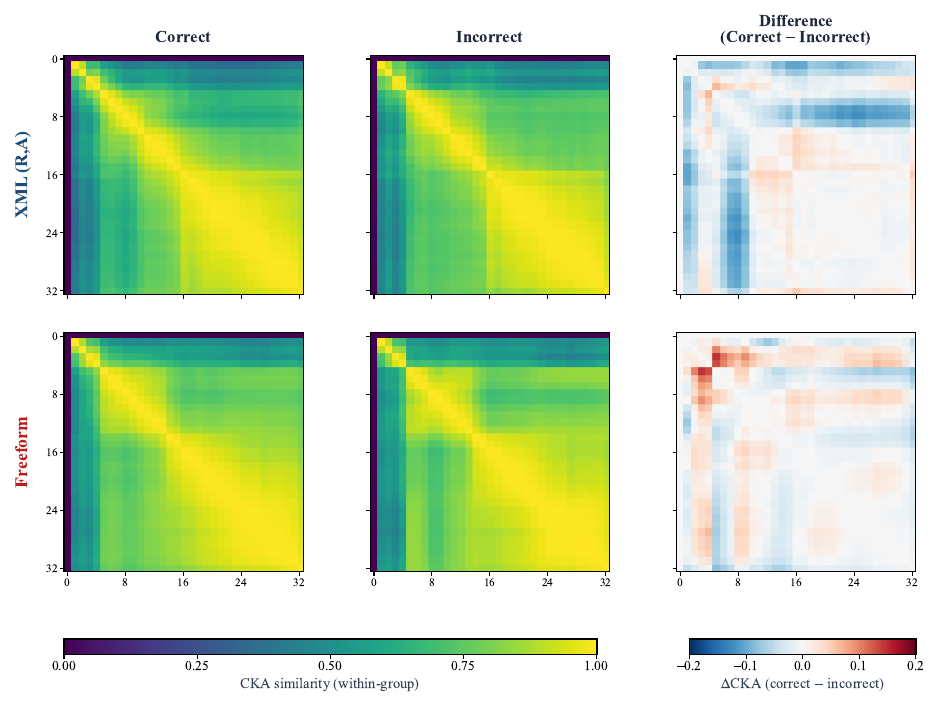}
          \caption{GSM8k}\label{fig:heatmap_gsm8k}
      \end{subfigure}\vspace{2pt}

      \begin{subfigure}{\textwidth}
          \includegraphics[width=\textwidth, height=0.421\textheight, keepaspectratio]{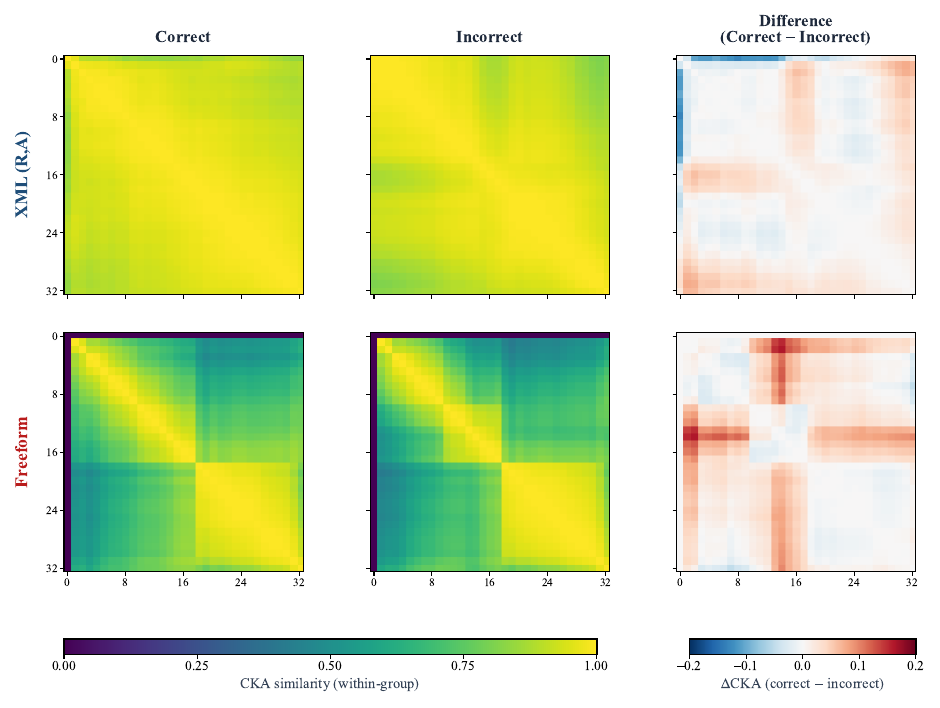}
          \caption{MMLU}\label{fig:heatmap_mmlu}
      \end{subfigure}\vspace{2pt}
      
      \caption{Llama-3.1-8B within-group CKA heatmaps for (a) GSM8K and (b) MMLU. Each panel shows
      XML\,(R,A) (top row) and Freeform (bottom row) across three views: correct
      answers only, incorrect answers only, and their difference
      (correct $-$ incorrect). See \S\ref{sec:app_cka_heatmaps} for construction and reading guide.}
      \label{fig:cka_heatmaps_gsm8k_mmlu}
  \end{figure*}

\begin{figure*}[p]  
      \centering
      \setlength{\abovecaptionskip}{4pt}
      \setlength{\belowcaptionskip}{0pt}

      \begin{subfigure}{\textwidth}
          \includegraphics[width=\textwidth, height=0.421\textheight,
  keepaspectratio]{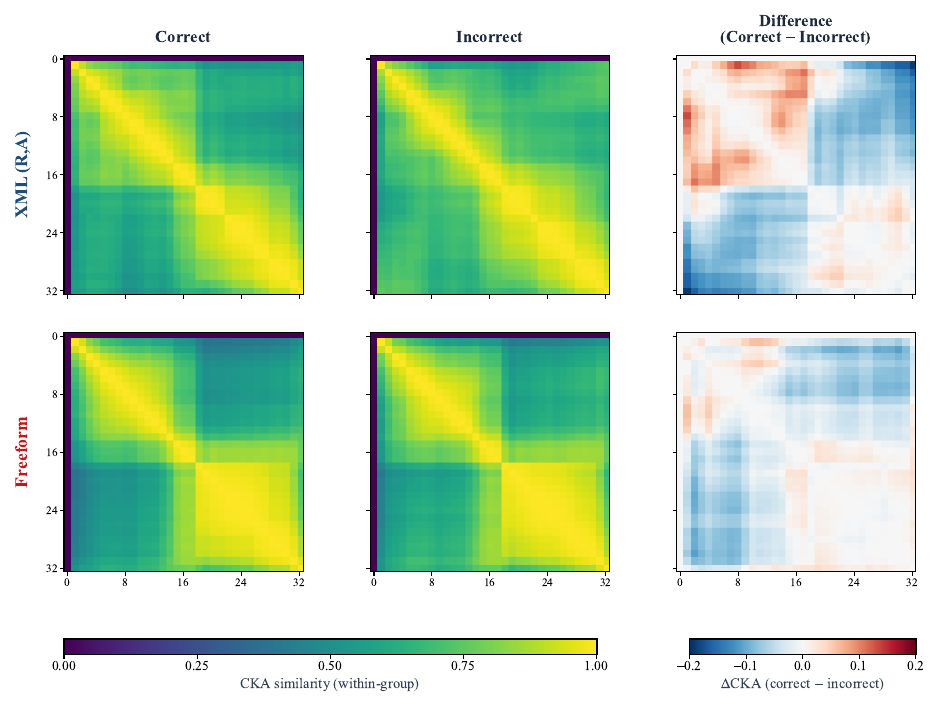}
          \caption{LogicBench}\label{fig:heatmap_logicbench}
      \end{subfigure}\vspace{2pt}

      \begin{subfigure}{\textwidth}
          \includegraphics[width=\textwidth, height=0.421\textheight, keepaspectratio]{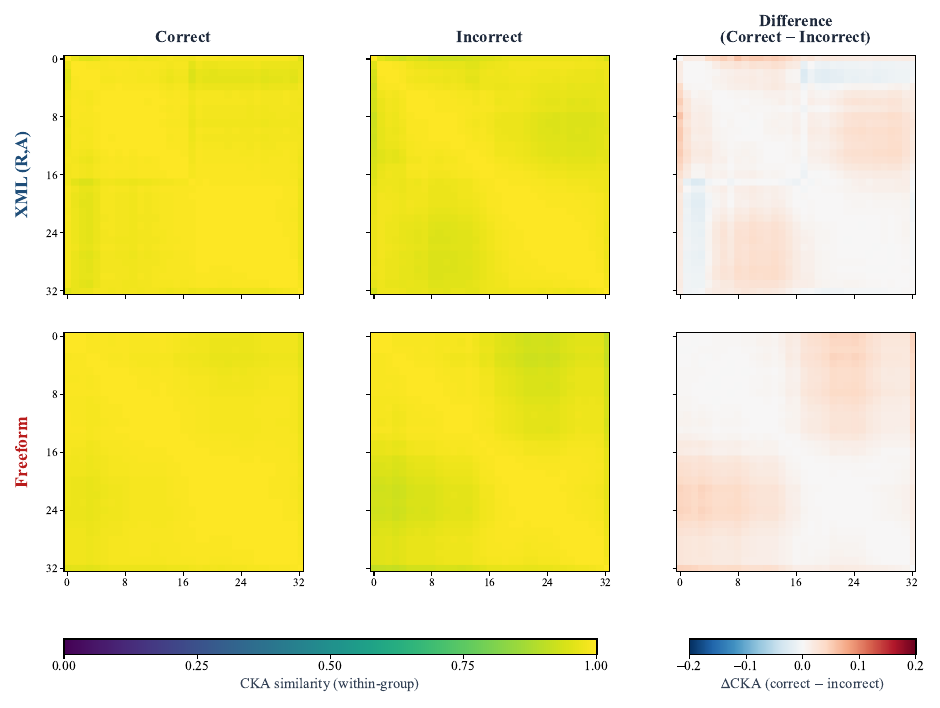}
          \caption{Spider}\label{fig:heatmap_spider}
      
      \end{subfigure}

      \caption{Llama-3.1-8B within-group CKA heatmaps for (a) LogicBench and (b) Spider. Same layout as Figure~\ref{fig:cka_heatmaps_gsm8k_mmlu}; see \S\ref{sec:app_cka_heatmaps}. Spider shows weak differentiation between outcome groups under both formats, consistent with the high-structure task pattern.}
      \label{fig:cka_heatmaps_logic_spider}
  \end{figure*}

\begin{figure*}[p]
    \centering
    \setlength{\abovecaptionskip}{4pt}
    \setlength{\belowcaptionskip}{0pt}

    \begin{subfigure}{\textwidth}
        \includegraphics[width=\textwidth, height=0.421\textheight, keepaspectratio]{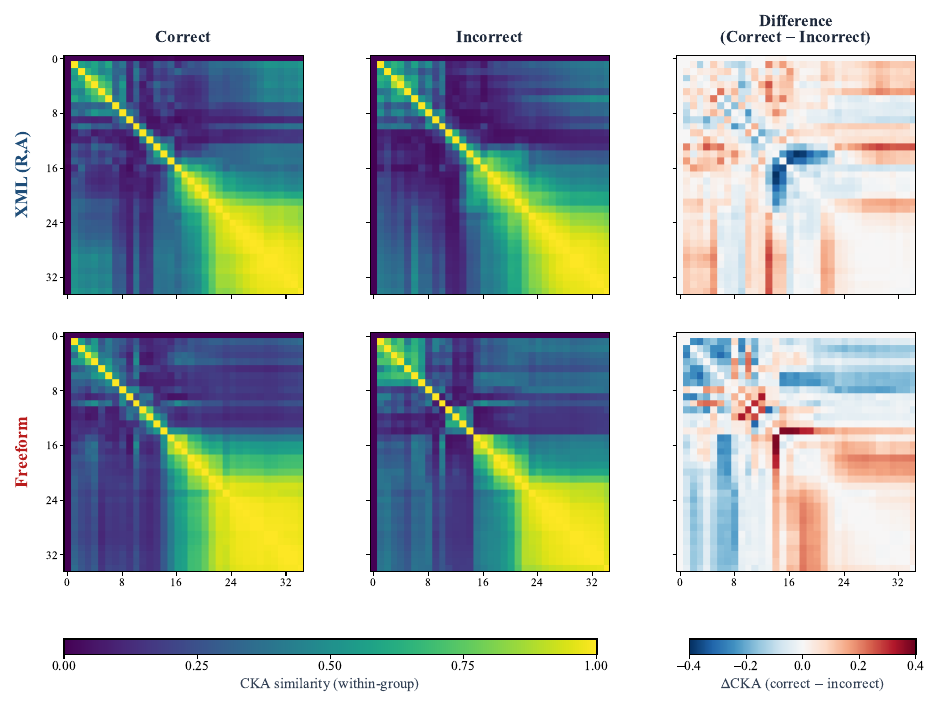}
        \caption{GSM8K}\label{fig:gemma_heatmap_gsm8k}
    \end{subfigure}\vspace{2pt}

    \begin{subfigure}{\textwidth}
        \includegraphics[width=\textwidth, height=0.421\textheight, keepaspectratio]{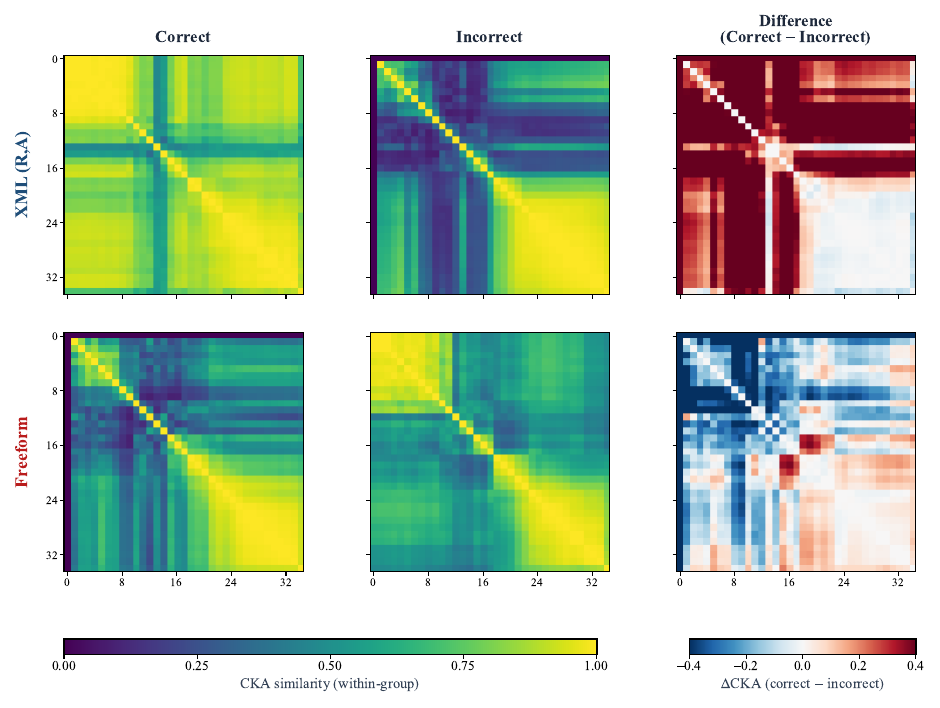}
        \caption{MMLU}\label{fig:gemma_heatmap_mmlu}
    \end{subfigure}\vspace{2pt}

    \caption{Gemma3-4B within-group CKA heatmaps for (a) GSM8K and (b) MMLU. Each panel shows XML\,(R,A) (top row) and Freeform (bottom row) across correct answers, incorrect answers, and their difference (correct $-$ incorrect). See \S\ref{sec:app_cka_heatmaps} for construction and reading guide.}
    \label{fig:gemma_heatmaps_gsm8k_mmlu}
\end{figure*}

\begin{figure*}[p]
    \centering
    \setlength{\abovecaptionskip}{4pt}
    \setlength{\belowcaptionskip}{0pt}

    \begin{subfigure}{\textwidth}
        \includegraphics[width=\textwidth, height=0.421\textheight, keepaspectratio]{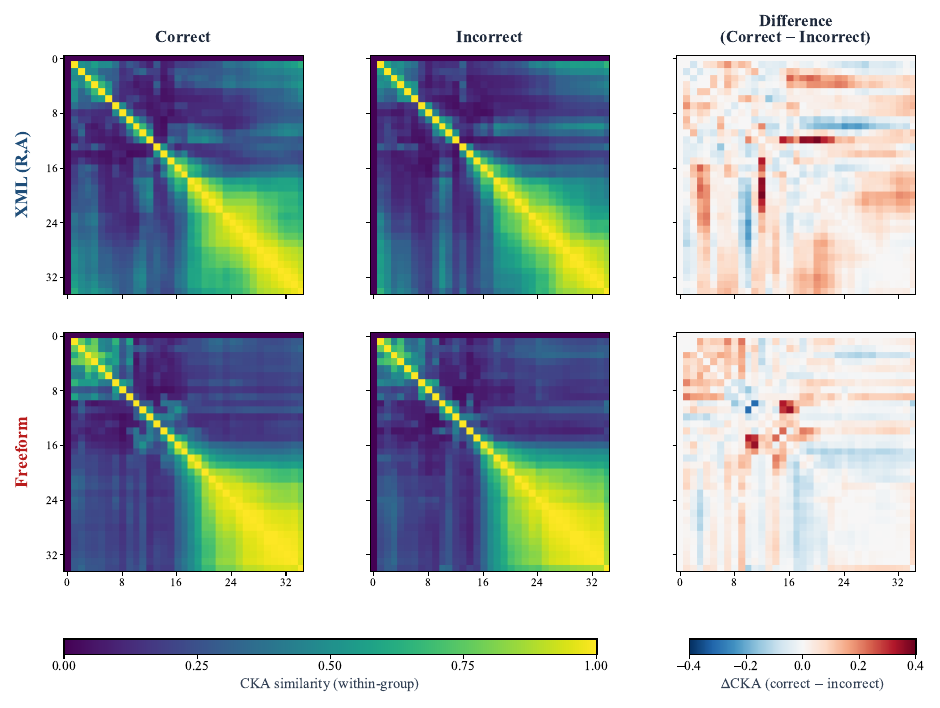}
        \caption{LogicBench}\label{fig:gemma_heatmap_logicbench}
    \end{subfigure}\vspace{2pt}

    \begin{subfigure}{\textwidth}
        \includegraphics[width=\textwidth, height=0.421\textheight, keepaspectratio]{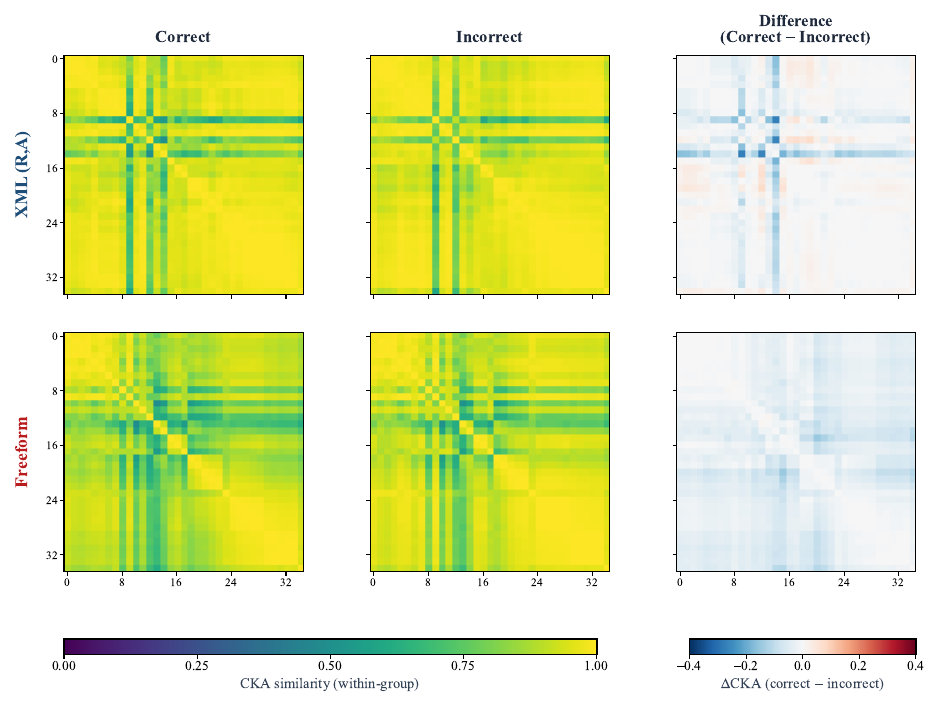}
        \caption{Spider}\label{fig:gemma_heatmap_spider}
    \end{subfigure}

    \caption{Gemma3-4B within-group CKA heatmaps for (a) LogicBench and (b) Spider. Same layout as Figure~\ref{fig:gemma_heatmaps_gsm8k_mmlu}; see \S\ref{sec:app_cka_heatmaps}. Spider shows weak differentiation between outcome groups under both formats, consistent with the high-structure task pattern.}
    \label{fig:gemma_heatmaps_logic_spider}
\end{figure*}

Within-group inter-layer CKA matrices are visualised in Figures~\ref{fig:cka_heatmaps_gsm8k_mmlu}--\ref{fig:cka_heatmaps_logic_spider} (Llama-3.1-8B) and Figures~\ref{fig:gemma_heatmaps_gsm8k_mmlu}--\ref{fig:gemma_heatmaps_logic_spider} (Gemma3-4B); see \S\ref{sec:app_cka_heatmaps} for layout and reading guide.

\subsection{Within-Group CKA Heatmaps: Layout and Reading Guide}
\label{sec:app_cka_heatmaps}

\paragraph{Matrix construction.}
For each (model, dataset, format) triple we partition the $n=200$ probe examples into two outcome groups: \emph{correct} ($C$, examples the model answered correctly) and \emph{incorrect} ($I$). For each group $g \in \{C, I\}$ we collect the last-token hidden state at every layer $\ell \in \{0, \dots, L-1\}$, yielding $\mathbf{X}^\ell_g \in \mathbb{R}^{n_g \times d}$. The within-group inter-layer CKA matrix $\mathbf{M}_g \in \mathbb{R}^{L \times L}$ then has entry
\[
\mathbf{M}_g[i, j] \;=\; \mathrm{CKA}\!\left(\mathbf{X}^i_g, \mathbf{X}^j_g\right),
\]
i.e.\ the linear CKA similarity between the population of layer-$i$ and layer-$j$ representations for examples in group $g$. This produces two $L \times L$ matrices per (format, dataset): one for the correct group and one for the incorrect group. We additionally show their difference, $\mathbf{D} = \mathbf{M}_C - \mathbf{M}_I$, which isolates representational-geometry differences attributable to outcome.

\paragraph{Panel layout.}
Each figure is a $2 \times 3$ grid:
\begin{itemize} 
    \item \textbf{Rows:} XML\,(R,A) (top), Freeform (bottom). Both rows use identically-sampled inputs; only the output format differs.
    \item \textbf{Columns:} (1) within-correct $\mathbf{M}_C$; (2) within-incorrect $\mathbf{M}_I$; (3) difference $\mathbf{D} = \mathbf{M}_C - \mathbf{M}_I$.
    \item \textbf{Colormaps:} columns 1--2 use \texttt{viridis} on $[0, 1]$ (CKA is in $[0, 1]$ by construction); column 3 uses a diverging \texttt{RdBu\_r} colormap symmetric about zero, with limits $\pm 0.20$ for Llama and $\pm 0.40$ for Gemma to accommodate the larger inter-layer variance of the smaller model.
    \item \textbf{Axes:} both axes index layer number, $0$ to $L-1$; the matrices are symmetric about the main diagonal by construction.
\end{itemize}

\paragraph{Reading guide.}
\begin{itemize} 
    \item \textbf{Block-diagonal structure} in the within-group panels (cols 1--2) indicates the model organises its forward pass into discrete representational stages: a contiguous run of layers that are mutually similar (a bright block on the diagonal) followed by an abrupt transition to a new regime. Three blocks is typical, corresponding to early (input encoding), middle (task computation), and late (output preparation) stages.
    \item \textbf{Off-diagonal bright patches} indicate two non-adjacent layers maintain similar representations -- a residual-stream signature. Conversely, dark off-diagonal patches mark regions where representations have substantially diverged.
    \item \textbf{Difference panel (col 3):} blue cells indicate the correct group has \emph{higher} within-layer similarity than the incorrect group at that $(i, j)$ location; red cells indicate the reverse. Saturation tracks magnitude.
\end{itemize}

\paragraph{What to attend to.}
\begin{itemize} 
    \item \textbf{Block-diagonal sharpness across rows.} Compare the top (XML R,A) and bottom (Freeform) rows in cols 1--2. On low-structure tasks (MMLU, LogicBench), XML\,(R,A) tends to produce a more crisply delineated three-block pattern in the correct group than Freeform does, consistent with structured prompting inducing cleaner stage-wise processing.
    \item \textbf{Asymmetry between correct and incorrect.} Within a single row, compare cols 1 vs.\ 2: a sharper block-diagonal pattern in the correct panel than in the incorrect panel suggests that correct answers are produced by more discretely-staged computation, while incorrect answers reflect noisier inter-layer transitions.
    \item \textbf{Difference panel as a summary.} A predominantly-blue difference panel indicates the correct group consistently has higher within-stage coherence than the incorrect group. A predominantly-red or near-white difference panel indicates outcome does not separate the representational geometry.
    \item \textbf{Where the difference concentrates.} A blue patch in the upper-left of $\mathbf{D}$ (early layers) suggests divergence begins early in the network; a mid-matrix concentration localises the differentiation to the task-computation stage.
    \item \textbf{Spider as the high-structure baseline.} On Spider, both formats and both outcome groups exhibit similar, less-pronounced block structure and a near-white difference panel -- the task's inherent SQL scaffolding already constrains representations, leaving little room for output format to modulate the internal geometry.
\end{itemize}

\newpage
\subsection{Linear Probe Setup}
\label{sec:app_probes}

We train a balanced linear probe (logistic regression with $\ell_2$ regularisation, $C=1.0$, 5-fold cross-validation) on the pre-answer hidden states at each layer to predict the model's own predicted label.
Hidden states are extracted from the last token of the reasoning field, immediately before the answer field begins, under both (R,A) and (A,R) orderings.

In both orderings, the probe reads the hidden state at the token immediately preceding the answer letter: the \texttt{>} closing the \texttt{<answer>} opening tag. This is offset $-1$ relative to the answer-letter token.

\begin{table}[h!]
\centering
\small
\label{tab:probe_position}
\begin{tabular}{lcll}
\toprule
\textbf{Ordering} & \textbf{Probe Token} & \textbf{Offset} & \textbf{Preceding Context} \\
\midrule
(R,A) & \texttt{>} in \texttt{<answer>} & $-1$ & Full reasoning block \\
(A,R) & \texttt{>} in \texttt{<answer>} & $-1$ & No reasoning; answer is the first output field \\
\bottomrule
\end{tabular}
\caption{Probe location relative to the answer token under different field orderings.}

\end{table}

We report balanced accuracy to account for class imbalance across answer options.
Probes are run on MMLU and LogicBench for {Llama-3.1-8B} and  {Gemma3-4B}; GSM8K is excluded as it requires a binary correct/incorrect probe rather than a label-identity probe, and Spider is excluded due to its open-ended SQL output space.
Hidden states for both models are extracted via forward pass on NVIDIA T4 GPUs.

\subsection{\textbf{Gemma3-4B}: Representational Results}
\label{sec:app_gemma_representations}
\begin{figure*}[h!]
    \centering
    \begin{subfigure}[t]{0.48\linewidth}
        \centering
        \includegraphics[width=\linewidth]{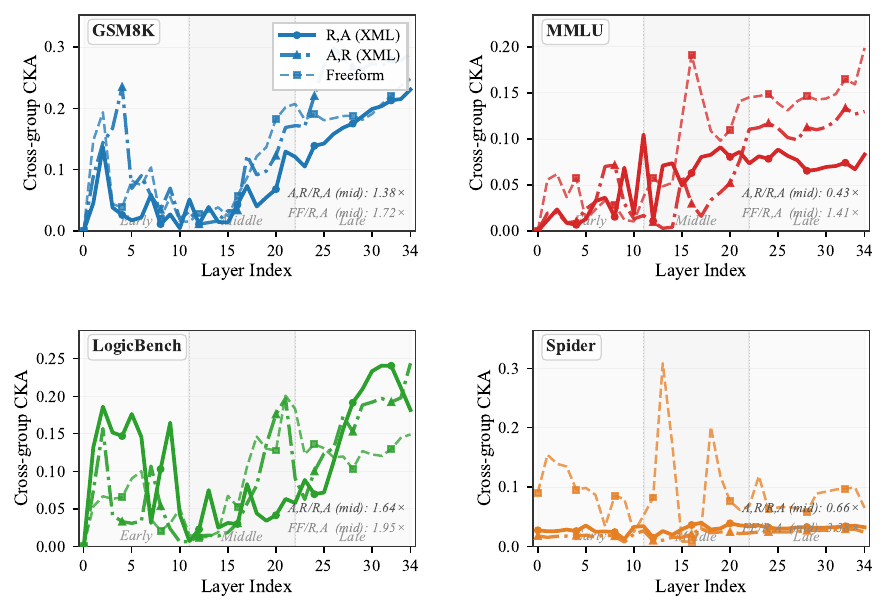}
        \caption{Cross-group CKA per layer}
        \label{fig:cka_gemma}
    \end{subfigure}
    \hfill
    \begin{subfigure}[t]{0.48\linewidth}
        \centering
        \includegraphics[width=\linewidth]{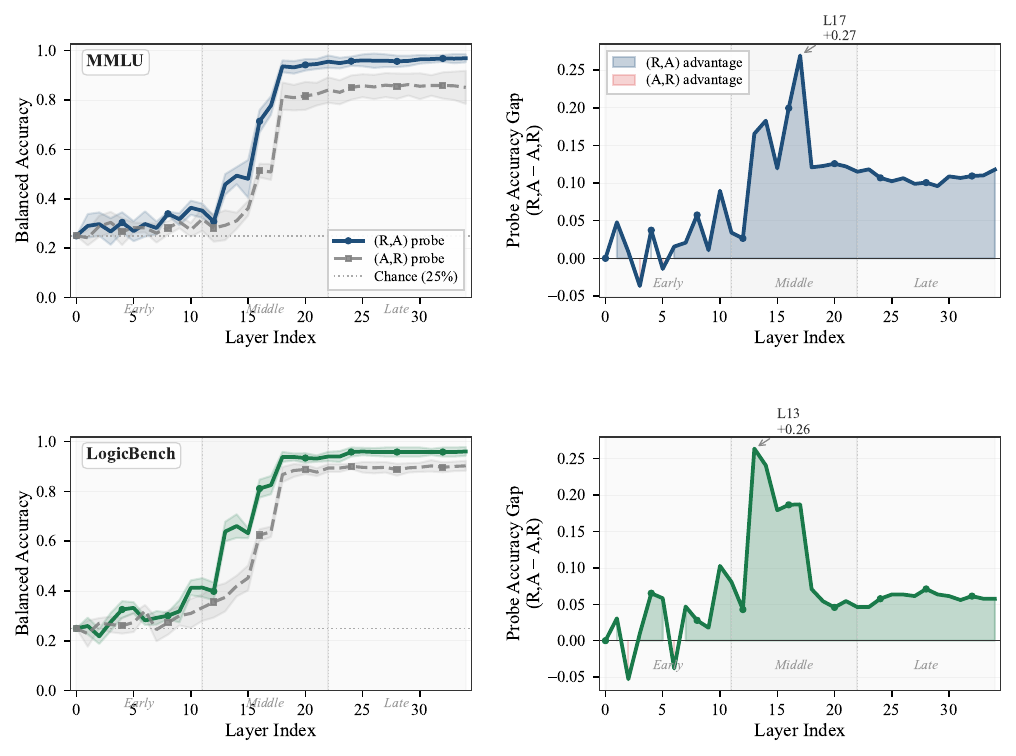}
        \caption{Linear probe balanced accuracy per layer}
        \label{fig:ap_gemma}
    \end{subfigure}
    \caption{Internal representations for \textbf{Gemma3-4B}, same layout as Figure~\ref{fig:representations}.
    \textbf{Left}: Cross-group CKA preserves the inverse-structure gradient of \textbf{Llama-3.1-8B}   MMLU and LogicBench show lower mid-layer CKA under XML\,(R,A) than freeform; Spider reverses the pattern at reduced overall magnitude.
    \textbf{Right}: Linear probe balanced accuracy on MMLU and LogicBench. The (R,A) advantage is directionally consistent with Llama but smaller in magnitude (peak gap $+0.27$ on MMLU at L17; $+0.26$ on LogicBench at L13), in line with Gemma's smaller behavioural gap under R,A in Table~\ref{tab:results}.}
    \label{fig:gemma_representations}
\end{figure*}

Figure~\ref{fig:gemma_representations} replicates the \textbf{Llama-3.1-8B} analysis of Section~\ref{sec:cka} for \textbf{Gemma3-4B} ($d=2560$, 35 layers).
The inverse-structure gradient in cross-group CKA is preserved: MMLU and LogicBench exhibit lower mid-layer CKA under XML\,(R,A) than freeform, while Spider again reverses the pattern   qualitatively identical to Llama but at reduced absolute magnitude, consistent with Gemma's smaller overall accuracy gap under (R,A) in Table~\ref{tab:results}.
Linear probes tell the same story: the (R,A) advantage in decoding answer identity from intermediate representations peaks at $+0.27$ on MMLU (L17) and $+0.26$ on LogicBench (L13), versus $+0.64$ and $+0.70$ for Llama   directionally aligned but attenuated, as expected for a model with fewer parameters and shallower representational depth.
The convergence across two architecturally distinct open-weight models strengthens the interpretation that mid-layer answer encoding under (R,A) is a general consequence of field ordering rather than a property of any single architecture.

\newpage
\section{Reasoning Length: Measurement and Results}
\label{sec:app_length}

Reasoning length is measured by extracting the reasoning field from each model response and counting words via \texttt{str.split()}, which closely tracks subword-token count under standard tokenizers.
For XML formats we extract the content of the \texttt{<reasoning>} tag; for JSON formats we parse the \texttt{"reasoning"} key, with a regex fallback for malformed outputs.
For Spider under JSON\,(A,R), \texttt{parsed\_reasoning} is empty in the primary data file and is extracted directly from raw response CSVs.
This yields 80 cells (5 models $\times$ 4 datasets $\times$ 4 format-orderings); each cell reports median word count and mean accuracy over all examples in that condition.

Tables~\ref{tab:token_distribution} and~\ref{tab:length_corr} report the full per-cell medians and Pearson correlations discussed in \S\ref{sec:length}.

\begin{table}[h!]
\centering\small
\setlength{\tabcolsep}{6pt}
\renewcommand{\arraystretch}{1.25}
\resizebox{0.475\textwidth}{!}{%
\begin{tabular}{@{}l rrrr@{}}
\toprule
\textbf{Dataset} & \textbf{XML (R,A)} & \textbf{XML (A,R)} & \textbf{JSON (R,A)} & \textbf{JSON (A,R)} \\
\midrule
  GSM8K      & 126 & 127 & \textbf{108} & 123 \\
  MMLU       & 121 &  99 &          121 & \textbf{33} \\
  LogicBench & 114 & \textbf{92} & 113 &  96 \\
  Spider     & \textbf{54} &  58 &  62 &  64 \\
\bottomrule
\end{tabular}
}
\caption{Median reasoning-field word count per format--ordering, aggregated over 5 models. \textbf{Bold} = shortest per dataset. JSON\,(A,R) collapses to 33 words on MMLU vs.\ 121 under XML\,(R,A); GSM8K is flat across all variants.}
\label{tab:token_distribution}
\end{table}

\begin{table}[h!]
\centering\small
\setlength{\tabcolsep}{8pt}
\renewcommand{\arraystretch}{1.25}
\resizebox{0.475\textwidth}{!}{%
\begin{tabular}{@{}lrr@{\hspace{1.5em}}lrr@{}}
\toprule
\multicolumn{3}{c}{\textbf{By dataset}} & \multicolumn{3}{c}{\textbf{By format--ordering}} \\
\cmidrule(lr){1-3}\cmidrule(lr){4-6}
\textbf{Dataset} & $r$ & $n$ & \textbf{Format} & $r$ & $n$ \\
\midrule
  GSM8K      & $-0.11$ & 20 & XML\,(R,A)  & $+0.42$ & 20 \\
  MMLU       & $+0.11$ & 20 & XML\,(A,R)  & $+0.28$ & 20 \\
  LogicBench & $+0.06$ & 20 & JSON\,(R,A) & $+0.25$ & 20 \\
  Spider     & $+0.16$ & 20 & JSON\,(A,R) & $-0.73$$^{**}$ & 20 \\
\addlinespace[2pt]\midrule
  Overall    & $+0.04$ & 80 &             &                 &    \\
\bottomrule
\end{tabular}
}
\caption{Pearson $r$ between median reasoning length and accuracy across 80 cells (5 models $\times$ 4 datasets $\times$ 4 format-orderings). $^{**}p{<}0.01$. Conditioning on dataset yields no signal; conditioning on format-ordering reveals JSON\,(A,R) as the sole significant predictor   and it is negative.}
\label{tab:length_corr}
\end{table}

\section{Prompt-Brittleness Ablation}
\label{sec:app_brittleness}

\paragraph{Motivation.}
LLM behaviour is known to be sensitive to surface-level prompt variations. Small changes in instructional phrasing, schema tag casing, or whitespace can shift accuracy by several points without changing semantics as emperically demonstrated by ~\cite{sclar2024quantifying, he2024doespromptformattingimpact}. A natural concern is whether the (R,A) vs.\ (A,R) accuracy gap reported in Section \ref{main_results} is similarly an artefact of the specific system prompts used. We address this concern with a focused robustness study.

\paragraph{Setup.}
We hold the model, dataset, and field ordering fixed and perturb only surface features of the system prompt. For each of the 5 formats (Free-form, JSON\,(A,R), JSON\,(R,A), XML\,(A,R), XML\,(R,A)), we construct three perturbed variants:

\begin{itemize} 
    \item \textbf{V1 Instructional phrasing.} Paraphrase the role and task sentence (``\color{blue} You are a math expert. Solve the problem step-by-step \normalcolor \ldots'' $\rightarrow$ `` \color{purple}Please solve the following mathematics problem carefully. Work through your reasoning \normalcolor \ldots''). Schema definition is byte-identical to baseline.
    \item \textbf{V2 Surface casing.} For XML, tag names become PascalCase (\color{blue}\texttt{<reasoning>} \normalcolor$\rightarrow$ \color{purple}\texttt{<Reasoning>}\normalcolor); for JSON, keys become PascalCase (\color{blue} \texttt{"answer"} \normalcolor $\rightarrow$ \color{purple}\texttt{"Answer"}\normalcolor); for Free-form, instruction-frame headers are uppercased. Instructional sentence is preserved.
    \item \textbf{V3 Whitespace.} For XML, blank lines around the schema are removed (compact form); for JSON, the schema definition is collapsed to a single line; for Free-form, additional blank lines are inserted around the separator. All wording and schema names are preserved.
\end{itemize}

We evaluate each (model, format, variant) cell on GSM8K dataset, across two models: Llama-3.1-8B (the model with the largest (A,R) collapse in the main results) and GPT-4o (a frontier API model). 

\begin{table}[h!]
\centering\small
\setlength{\tabcolsep}{4pt}
\renewcommand{\arraystretch}{1.15}
\begin{tabular}{@{}llcccc@{}}
\toprule
\textbf{Model} & \textbf{Format} & \textbf{Base} & \textbf{V1} & \textbf{V2} & \textbf{V3} \\
\midrule
\multirow{5}{*}{Llama-3.1-8B}
    & Free-form     & 77.5 & 81.0 & 78.5 & 80.0 \\
    & JSON\,(A,R)   & 32.0 & 35.0 & 34.5 & 36.0 \\
    & JSON\,(R,A)   & 81.5 & 86.0 & 74.5 & 77.0 \\
    & XML\,(A,R)    & 10.5 & 11.5 & 11.0 & 11.5 \\
    & XML\,(R,A)    & 84.0 & 87.5 & 82.0 & 82.0 \\
\midrule
\multirow{5}{*}{GPT-4o}
    & Free-form     & 90.0 & 86.0 & 90.0 & 86.5 \\
    & JSON\,(A,R)   & 59.0 & 56.5 & 55.5 & 56.5 \\
    & JSON\,(R,A)   & 96.0 & 98.0 & 96.0 & 97.0 \\
    & XML\,(A,R)    & 57.0 & 53.5 & 55.5 & 53.5 \\
    & XML\,(R,A)    & 96.0 & 97.0 & 98.0 & 98.5 \\
\bottomrule
\end{tabular}
\caption{GSM8K accuracy (\%) dataset under three prompt-perturbation variants. \textbf{Base} = original prompt used in the main results. Within each (model, format-family) pair, the (R,A) row consistently exceeds the (A,R) row across all four conditions.}
\label{tab:brittleness}
\end{table}

\begin{figure}[h!]
  \centering
  \includegraphics[width=\columnwidth]{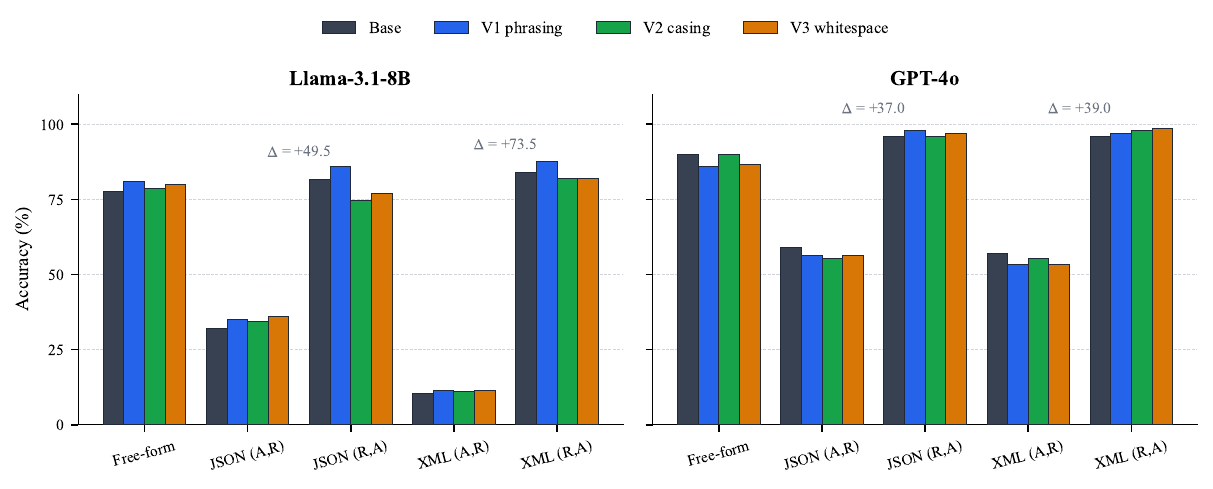}
  \caption{GSM8K accuracy under prompt perturbations. Each group of bars is one (model, format) cell; within-group order is baseline, V1, V2, V3. The (R,A) $-$ (A,R) gap (annotated $\Delta$ over each XML / JSON family) is preserved across all perturbations: \textbf{the sign of the field-ordering effect is invariant to surface prompt changes}, even though absolute accuracy shifts by a few points.}
  \label{fig:brittleness}
\end{figure}

\paragraph{Findings.}
\textbf{(1)~Sign of the gap is preserved.}
For every (model, format-family, variant) cell, accuracy under (R,A) exceeds accuracy under (A,R); no perturbation inverts the ordering. The (R,A) $-$ (A,R) gap on Llama XML is +73.5\,pts (range across variants: +70.5 to +76.0); on GPT-4o XML it is +39.0\,pts (range +39.0 to +45.0); analogous magnitudes hold for JSON. The specific concern that field ordering's effect could disappear under slight prompt changes is empirically rejected.

\textbf{(2)~Absolute accuracy shifts are small relative to the ordering effect.}
The largest within-cell spread across $\{$Base, V1, V2, V3$\}$ is 11.5\,pts (Llama JSON\,(R,A); Table~\ref{tab:brittleness}); most cells move by under 5\,pts. By contrast, the \emph{smallest} (R,A) $-$ (A,R) gap across all $(\text{model}, \text{family}, \text{variant})$ combinations is +37\,pts (GPT-4o JSON baseline).

\end{document}